\documentclass[acmtocl,acmnow]{acmtrans2m}

\newdef{definition}[theorem]{Definition}
\newdef{remark}[theorem]{Remark}

\usepackage{graphicx}
\usepackage{url}
\usepackage{comment}
\usepackage{color}
\usepackage{amsmath}
\usepackage{multirow}

\title{Detecting and Extracting Events from Text Documents}
            
\author{JUGAL KALITA, University of Colorado, Colorado Springs}
            
\begin{abstract} 
Events of various kinds are mentioned and discussed in text documents, whether they are books, news articles, blogs or microblog feeds. The paper starts by giving an overview of how events are treated in linguistics and philosophy. We follow this discussion by surveying how events and associated information are handled in computationally. In particular, we look at how textual documents can be mined to extract events and ancillary information. These days, it is mostly through the application of various machine learning techniques. We also discuss applications of event detection and extraction systems, particularly in summarization, in the medical domain and in the context of Twitter posts. We end the paper with a discussion of challenges and future directions. 
\end{abstract}
            
\category{...}{...}{...}
            
\terms{...} 
            
            
\begin{document}
            
\begin{bottomstuff} 
Jugal Kalita is a professor in the Department of Computer Science, University of Colorado, Colorado Springs, CO 80918 USA . 
\end{bottomstuff}
            
\maketitle

\section{Introduction}
Among the several senses that The Oxford English Dictionary\footnote{\url{http://www/oed.com}}, the most venerable dictionary of English,  provides for the word {\em event}, are the following. 
\vspace*{-2pt}
\begin{itemize}
\setlength{\itemsep}{-1pt}
\item [1a] {\em the (actual or contemplated) fact of anything happening; the occurrence {\em of}}.
\item [2a] {\em anything that happens, or is contemplated as happening; an incident, occurrence}.
\item [2d] {\em (In modern use, chiefly restricted to) occurrences of some importance}. 
\end{itemize}
\vspace*{-2pt}
Although an event may refer to anything that happens, we are usually interested in occurrences that are of some importance. We want to extract such events from textual documents. In order to extract important events or events of a specific type, it is likely that we have to identify all events in a document to start with. 

Consider the first paragraphs of the article on the {\em Battle of Fredericksburg} in the English Wikipedia, accessed on May 5, 2012. We have highlighted the ``events" in the paragraph. 

\begin{quote}
{\em 
The Battle of Fredericksburg was {\bf  fought} December 11--15, 1862, in and around Fredericksburg, Virginia, between General Robert E. Lee's Confederate Army of Northern Virginia and the Union Army of the Potomac, {\bf commanded} by Maj. Gen. Ambrose E. Burnside. The Union army's futile frontal {\bf  assaults} on December 13 against {\bf entrenched} Confederate defenders on the heights behind the city is {\bf remembered} as one of the most one-sided {\bf battles} of the American Civil War, with Union {\bf casualties} more than twice as heavy as those {\bf suffered} by the Confederates.
}
\end{quote}

The paragraph contains two fairly long sentences with several ``events", mentioned using the following words: {\em fought}, {\em commanded}, {\em assaults}, {\em entrenched}, {\em remembered}, {\em casualties} and {\em suffered}. Some of these ``events" are described in terms of verbs whereas the others are in terms of nouns. 
Here {\em fought}, {\em commanded}, {\em assaults}, {\em battles} definitely seem to be ``events" that have durations or are {\em  durative}.  {\em Entrenched} seems to talk about a state, whereas it is possible that {\em suffered} talks about something {\em punctual} (i.e., takes a moment or point of time) or can be {\em durative} (i.e., takes a longer period of time) as well. The act of remembering by an individual is usually considered to happen momentarily, i.e., forgotten things come back to mind at an instant of time. But, in this paragraph it is given in passive voice and hence, it is unclear who the actor is, possibly a lot different people at different points of time. 

Thus, depending on who is asked, the ``events'' picked out may be slightly different, but the essence is that there are several events mentioned in the paragraph and the objective in event extraction is to extract as many of them as possible in an automated fashion. For example, someone may not pick out {\em remembered} as an event that took place. Some others may not want to say that {\em entrenched} is an event. In addition, if one is asked to pick an important event, responses may vary from person to person. Finally, if one is asked to summarize the paragraph, depending on the person asked the summary may vary. A summary prepared by the author of this article is given below.
\begin{quote}
{\em
The Battle of Fredericksburg, fought December 11-12, 1862, was one of the most one-sided battles of the American Civil War, with heavy Union casualties.
}
\end{quote}
Obviously, there are many other possibilities for summarization. However, the idea is that identification of events and their participants may play a significant role in summarizing a document. 

This paper discusses the extraction of events and their attributes from unstructured English text. 
It is an  survey of research in extracting event descriptions from textual documents. In addition, we discuss how the idea of event extraction can be used in application domains such as summarization of a document. We also discuss application of event extraction in the biomedical domain and in the context of Twitter messages. 

The rest of the paper is organized in the following manner. Section~\ref{section:linguisticEvents} provides a description of research in linguistics and philosophy. The author believes that such a background, at least at a basic level,  is necessary to understand and develop the approaches and algorithms for automatic computational detection and extraction of events and their participants from textual documents. Section~\ref{section:eventExtraction} discusses approaches used in extracting events from textual documents. Most approaches these days use machine learning techniques. 

\section{Events in Linguistics and Philosophy}
\label{section:linguisticEvents}
Real world events are things that take place or happen. In this section, we present an overview of how real events are represented in terms of language. In particular, we discuss classification of events and features necessary for such classification. We follow this by presenting the preferred way among philosophers  to represent events in terms of logic. We bring this section to an end by presenting some of the structures  ascribed to events by linguists or philosophers working at an abstract level. 

The reason for the inclusion of this section in the paper  is  to set the context for the discussions in the following sections on the practical task of extracting events. Practical systems do not usually follow linguistic niceties although they draw inspiration from linguistics or philosophy.

\subsection{Classifying Events}
There have been many attempts at classifying linguistic events. Below, we briefly discuss a few. 
The primary focus when linguists discuss events is on the verb present in a sentence. Nouns, adjectives and other elements present in a sentence provide arguments for the verb.  

Aristotle (as presented in \cite{Barnes1984}) classified verbs that denote something happening into three classes: {\em actuality}, {\em movement} and {\em action}. An actuality represents the existence of a thing or things; this is called {\em state} by others (e.g., \cite{Rosen1999}). An examples of actuality can be seen in the sentence {\em Jon is ill}. A movement is an incomplete process or  something that  takes time but doesn't have an inherent end.  An example of movement is seen in the sentence {\em  Jon is running}. An action is something that takes time and has an inherent end. An example of an action is present in the sentence {\em Jon is building a house}. In other words, Aristotle distinguished between states and events and then events.

 \cite{Kenny2003}  lists verbs that belong to the three Aristotelian classes and develops membership criteria for the classes. Kenny renamed the classes as {\em states}, {\em activities} (actions without inherent end) and {\em performances}  (actions with inherent ends). Kenny's membership criteria are based on semantic entailments about whether the event can be considered to have taken place when it is still in progress. For example, during any point when we say {\em Jon is running}, we can consider that the activity of running has taken place. In other words {\em Jon is running}  entails {\em Jon has run}.  Thus, {\em run} is an {\em activity}. In contrast, when we say {\em Joh is taking the final}, we cannot say that Jon has taken the final. In other words, the first does not entail the second. Thus, the main difference between an activity and a performance is what is called {\em delimitation}. A delimited event has a natural end. 

 \cite{Vendler1967} developed a 4-way classification scheme for linguistic events and  \cite{Dowty1979} developed a set of  criteria for membership in the classes. The classes enumerated by Dowty are: {\em states}, {\em activities}, {\em achievements} and {\em accomplishments}. The definitions are given below.

\begin{itemize}
\item {\em Activities}: Events that take place over a certain period of time, but do not necessarily have a fixed termination point. Examples; {\em Jon walked for an hour}, and {\em Jon is driving the car}. 
\item {\em Accomplishments}: Events that happen over a certain period of time and then end. Examples: {\em Jon built a house in a month}, and {\em Jon is taking the final}. 
\item {\em Achievements}: These are events that occur instantaneously and lack continuous tenses.  Examples: {\em Jon 
finished the final in 45 minutes} and {\em The vase broke}. 
\item {\em States}: These are non-actions that hold for a certain period of time, but lack continuous tenses. Examples: {\em Jon knows the answer} and {\em Jon likes Mary}.
\end{itemize}

\cite{Smith1997} adopts the same classification as Vendler and Dowty, but divides {\em achievements} into two classes. The first one is still called {\em achievements}, but the second one is called {\em semelfactives}. In this new scheme, achievements are instantaneous (that is, the beginning of the event is the same as its end)  culminating events, but semelfactives are events with no duration that result in no change of state. An example of a semelfactive is: {\em Jon knocked on the door}. 

Table~\ref{table:eventClassification} presents the nomenclatures introduced by various linguists in one place. There are many variations of the schemes given here, although we do not discuss them in this paper. 
\begin{table}
\begin{center}
\begin{tabular}{| l | l |} \hline
Linguist & Nomenclature used \\ \hline
Aristotle & Actuality, Movement, Action \\ \hline
Kenny & State, Activity, Performance \\ \hline
Dowty & State, Activity, Accomplishment, Achievement \\ \hline
Smith & State, Semelfactive, Activity, Accomplishment, Achievement\\ \hline
\end{tabular}
\caption{Nomenclatures used by linguists to classify events}
\label{table:eventClassification}
\end{center}
\end{table}

In the early work on event classification, Aristotle, Vendler and others assume that what needs to be classified is the verb. However, many have concluded that it is impossible to classify a verb into a specific class.  It is more appropriate to say that a clause containing an event has a class, and the classification of such a clause depends not only upon the verb, but also on other material present in the clause \cite{Rosen1996,Dowty1979,Dowty1991,Ritter1996}. In other words, the classification must be compositional or must depend on various features of the clause, not exclusively verb-based. There is also substantial evidence that sentence material other than the verb can change the overall event type. For example, addition of a direct object can change an activity to an accomplishment \cite{Rosen1999}, as in the following examples.
\begin{itemize}
\item {\em Bill ran for five minutes/*in five minutes}: activity
\item {\em Bill ran the mile *for 5 minutes/in 5 minutes}: accomplishment
\end{itemize}

\subsection{Parameters of Event Classes}
Many authors in linguistics have delved deeper into the nature of event classes and have tried to come up with features or characteristics that can be used to identify whether something (verb or a clause)  belongs to a particular event class or not. These features or characteristics are necessary to describe the {\em structure of events} in a theoretical sense. Description of event structure usually refers to the actual words used (lexical features or characteristics) and also the structure of clause or sentence (syntactic features or characteristics). Identification of such features may be described as finding parameters of event types or {\em parameterization of event types}. 

A lot of the work on parameterization of event types/classes use the classes espoused by Vendler. These include \cite{Verkuyl1996,Carlson1981,Moens1987,Hoeksema1983,Mourelatos1978,TerMeulen1983,TerMeulen1997} and others. We will only briefly touch upon such work in this paper. Our objective is to impress upon the reader that identification of features of event classes is considered an important task by linguists. 

For example, \cite{Verkuyl1996} describes Vendler's classes with two binary features or parameters: {\em continuousness}: whether an event has duration, and {\em boundedness}: whether an event has a (natural) terminal point or endpoint. Using these two features, the four Vendler classes can be parameterized as follows.

\begin{itemize}
\item [{\em state}]: -bounded, -continuous
\item [{\em activity}]: -bounded, +continuous
\item [{\em achievement}]: +bounded, -continuous
\item [{\em accomplishment}]: +bounded, +continuous
\end{itemize}

\cite{Hoeksema1983,Mourelatos1978} introduce the notion of {\em countability} while discussing event classes. This is similar to the mass-count opposition in nouns. Terminating events can be counted, but non-terminating processes cannot. Hoeksema introduces two binary features: {\em count} and {\em duration} to obtain Vendler's classes as seen below. The feature duration refers to whether the event takes place over time. 
\begin{itemize}
\item [{\em state}]: -count, -duration
\item [{\em activity}]: -count, +duration
\item [{\em achievement}]: +count, -duration
\item [{\em accomplishment}]: +count, +duration
\end{itemize}

\cite{Moens1987} refines Vendler's classes by adding a class much like Smith's semelfactives \cite{Smith1997}. He suggests that, in addition to states, there are four event types: {\em culmination}, {\em culminated process}, {\em point}, and {\em process}. He uses two binary features or parameters: {\em consequence} identifying termination or culmination, and {\em atomic} or non-atomic (which Moens called {\em extended}). Atomic is also called {\em momentous} or {\em pointed}.  Moen's classification is given below, along with the features and examples.
\begin{itemize}
\item [{\em  culmination}]: +consequence, +atomic  (examples: {\em recognize}, {\em win the race})
\item [{\em culminated process}]: +consequence, -atomic (examples: {\em build a house})
\item [{\em point}]: -consequence, +atomic (example: {\em hiccup}, {\em tap}, {\em wink})
\item [{\em process}]: -consequence, -atomic (example: {\em run}, {\em swim}, {\em play the piano})
\item [{\em state}]: (examples: {\em understand}, {\em love}, {\em resemble})
\end{itemize}
Moens also claims that {\em culminated process} is an event class whose members are made up of smaller atomic units. In particular, a culminated process is a {\em process} with a consequent state. This insight that events can be decomposed into sub-events was used later by others working on the lexical analysis of events e.g., \cite{Pustejovsky1991a,Pustejovsky1991b}. Others such as \cite{vanVoorst1988,Grimshaw1990,Tenny1994} have claimed that arguments of verbs are related to sub-events. 

We summarize the various features that linguists have used to classify events in Table~\ref{table:eventFeatures}. Of course, we do not discuss many other proposals for features in this brief discussion. 
\begin{table}
\begin{center}
\begin{tabular}{| l | l |}\hline
Linguist & Event features identified \\ \hline
Verkuyl & bounded, continuous\\ \hline
Hoeksema & count, duration\\ \hline
Moens & consequence, atomic \\ \hline
\end{tabular}
\caption{Features used by linguists to classify events}
\label{table:eventFeatures}
\end{center}
\end{table}
Classification of events and their parameterization of verbs or predicates (or clauses) are only the first steps in developing a deeper linguistic understanding of events. In particular, in order to understand the linguistic representation of events, linguists need to go beyond classification schemes.

\subsection{Events in Logical Representation of Semantics}
Mathematical logic is used to represent the semantics of language. In particular, we use logic to represent the meaning of single sentences. Early work on events, e.g., Panini (as discussed by \cite{Parsons1990} and \cite{HamiltonPlato1961}) stated that language encodes two kinds of information--{\em actions} and {\em non-actions}. Verbs represent actions and nouns represent non-actions or things. 

\cite{Davidson1967} proposes that one needs an event variable $e$ to represent events in mathematical logic.  This variable $e$ is used to represent relations represented by the event denoted by the verb and other constituents in the sentence, such as modifiers. Davidson claims that logically speaking, events are like things in that they can be represented by a variable and this variable can be modified and quantified. A question that arises is: how many arguments should an event predicate (in logic) take \cite{Kenny2003}? Just like nominal modifiers modify nouns, event  modifiers can modify event predicates.  An event predicate can take any number of  modifiers just like noun (nominal) modifiers. Examples of event modifiers are: time, place, manner and instrument. Davidson proposed that an event predicate may take one or more required arguments (is this true?) and any number of adjuncts or optional modifiers. 
Consider the following examples from \cite{Davidson1967}. The English sentence and the corresponding logical representation or {\em logical form} is given for each example. 
\begin{itemize}
\item [a.] {\em John buttered the toast.} \\ 
	$\exists e \; buttered (Jones, the\_toast, e)$
\item [b.] {\em John buttered the toast slowly.}\\ 
	$\exists e \; buttered (Jones, the\_toast, e) \wedge slowly (e)$
\item [c.] {\em John buttered the toast slowly, in the bathroom.}\\
	 $\exists e \; buttered (Jones, the\_toast, e) \wedge slowly (e) \wedge in\_the\_bathroom (e)$
\item [d.] {\em John buttered the toast slowly, in the bathroom, with a knife.}\\
	 $\exists e \; buttered (Jones, the\_toast, e) \wedge slowly (e) \wedge in\_the\_bathroom (e) \wedge with\_a\_knife (e)$
\item [e.] {\em John buttered the toast slowly, in the bathroom, with a knife, at midnight.}\\
	 $\exists e \; buttered (Jones, the\_toast, e) \wedge slowly (e) \wedge in\_the\_bathroom (e) \wedge with\_a\_knife (e) \wedge at\_midnight (e)$
\end{itemize}
Thus we can see that Davidson's approach places the event variable $e$ in the main predicate of a clause and distributes it among the modifiers of the clause in logical representation. In writing the meaning in Davidsonian logic, the author creates predicates such as $the\_toast$ and $in\_the\_bathroom$, just for illustration, without going into details.

Davidsonian representation allows events to be represented in logic (logical semantics) without requiring verbs to have multiple arities, i.e., without taking different arguments in different situations. Because the event is represented as a variable, the event variable $e$ can be included in the representation of logical meaning of each modifier or adjunct.  Another benefit is that using Davidson's representation, one can analyze events represented syntactically as nouns (nominals) or verbs \cite{Parsons1990}. For example, one can refer to an event using the verb {\em to burn} or the noun {\em a burn}. Parsons also observes that using a variable to represent an event allows quantification over events the same way quantification applies to things. The following examples are from \cite{Parsons1990}.
\begin{itemize}
\item [a.] {\em In every burning, oxygen is consumed}.\\
	$\forall e \; burning (e) \rightarrow  \exists e' (consuming (e') \wedge object (e, oxygen) \wedge in (e,e')$
\item [b.] {\em Agatha burned the wood}.\\
	$\exists e \; burning (e) \wedge subject (e, Agatha) \wedge object (e, wood)$
\item [c.] {\em Oxygen was consumed}.\\
	$\exists e' \; consuming (e') \wedge object (e', oxygen)$
\end{itemize}
We do not go into details of containment of events as expressed by $in$ in the first example above, and also the representation of passives as in the third example above. In these three examples, the author uses predicates such as {\em object} and {\em subject} which represent more fine-grained relationship with the main predicate (corresponding to the verb usually) than the examples earlier. 
Extending this work, \cite{Parsons1990,Higginbotham1985,Vlach1981} have demonstrated that using Davidson's $e$ variable allows one to express tense dependency between perception verbs and their infinitival compliments in a natural way. 

 \cite{Parsons1990} extends Davidson's approach to logical representation by adding an extra term corresponding to the event type of the predicate. He distinguishes between two types of eventualities: eventualities that culminate called {\em Cul} containing achievements and accomplishments, and those that do not, called {\em Hold} containing states and activities. 
\begin{itemize}
\item [a.] {\em John buttered the toast}.\\
	$\exists e \; buttering (e) \wedge agent (e, Jones) \wedge theme (e, toast) \wedge
			(\exists t \; (t <now \wedge Cul (e,t))$
\item [b.] {\em Mary knows Fred.}\\
	$\exists e \; knowing (e) \wedge experiencer (e, Mary) \wedge theme (e, Fred) \wedge Hold (e, now))$
\end{itemize}
In the logical representation in these examples, the author uses predicates such as {\em theme}, {\em agent} and {\em experiencer} which are usually are called {\em cases} in linguistics \cite{Fillmore1977}. In addition, the author uses a variable $t$ to express time. $now$ is a special indexical variable. We do not give detailed discussions of these fine points here. 

\cite{Hobbs1985} also proposes a logical form based on Davidson's approach. The main motivation behind Hobb's approach is to be able to produce satisfactory semantic representation when an event is expressed as a noun,  or when we want to express the meaning of tenses, modalities, and adverbial modifiers. He also explains how so-called {\em opaque adverbials} like {\em almost} in the sentence, {\em John is almost a man.} can be represented by the Davidsonian approach, which Hobbs extends. He also shows how the ambiguity between {\em de re} and {\em de dicto} meanings of sentences \cite{Quine1956} that discuss beliefs  can be explained by his approach to logical form representation of sentences. The representation by Hobbs is quite similar to other such representations based on Davidson, although there are some fine points of differences, that we do not discuss here. From a practical point of view, several research efforts in computational linguistics have adopted Hobb's logical form, and one such recent approach is  by \cite{Rudinger2014} who attempt to map Stanford dependency parses \cite{deMarneffe2006} into  Hobbsian logical form, and discover that sometimes it is possible to do so, but in other cases the mapping requires semantic information that is not present in such dependencies indentified by the Stanford parser. 

\subsection{ Event structure}
Early efforts at identification of event structure in linguistics was usually limited to explaining essential grammatical phenomena. However, others later proposed complex structures that go beyond  simple structures such as Davidson's approach of representing an event by a single logical variable and its components by additional predicates.   
Understanding the structure of an event entails (i) understanding the argument structure of the word (or, phrase) used to express the event in surface form, (ii) understanding the components in the conceptual or semantic description of an event, and (iii) understanding the relation or mapping between syntactic realization of an event and its conceptual components. 
In fact, analysis of argument structure includes all three steps and  requires finding the relation between meaning of a verb (or a clause) and the syntactic realization of arguments. \cite{Grimshaw1990,Williams1981} introduce argument structure as a distinct level of representation in linguistics.  Other prominent representations proposed include  f-structures \cite{Bresnan1982}, linear event structures \cite{vanVoorst1988}, lexical conceptual structures (LCS) \cite{Jackendoff1985,Rappaport1988} and two related structures: event structures and qualia structures for arguments \cite{Pustejovsky1991b}. 

There are two sides to event structure: syntactic and semantic. When specifying event structure, at the semantic level,  the description must be richer than semantic role descriptions \cite{Gruber1965,Fillmore1968}.  \cite{Levin1985} argues that named roles or thematic roles are too coarse-grained to provide useful semantic interpretation of a sentence. It is also necessary to capture semantic distinctions in a much more fine-grained manner compared to prior theories of \cite{Katz1963,Wilks1975,Quillian1968}. 
***A sentence or two on these theories***
By this time it was clear that  sophisticated approaches to specifying event structure must build upon the rich taxonomy of verb classes \cite{Levin1985} and descriptive vocabulary work \cite{Talmy1985} and \cite{Jackendoff1985}.

\subsubsection{Lexicalization Patterns by Talmy}
\cite{Talmy1985} discusses the systematic relations in language between meaning and surface expression. 
In particular, Talmy focuses on verbs and in particular, verbs that describe motion or location. He sketches a ``motion" event in order to explore issues in lexicalization. The basic motion event consists of one object called {\em figure} moving or located with respect to another object called the {\em referent} or the {\em ground}. The motion event has additional components such as {\em path} and  {\em motion},  {\em manner} and {\em cause}. Talmy gives examples of cases where the verb at once can express, in addition to the action or motion,  one or more of  figure, path,  manner or cause.  If a sematic component such as manner or cause is expressed directly by the verb, it is called conflation of manner (or cause) into the verb. 
  Some verbs incorporate aspect, which  represents the ``pattern of distribution of at ion though time."  In some languages, verbs can incorporate personation as well. Personation is a specification of the person involved, e.g., self or non-self. Some verbs incorporate what is called {\em valence}, where in conceptualizing an event that involves several different entities in distinct roles, a verb is able to direct greater attention to some one of these entities that to the others, or perhaps adopt a specific perspective. 
Sometimes, semantic components are not incorporated into the verb, but are expressed through what Talmy calls {\em satellites}. A satellite is an immediate constituent of the verb root other than inflections, auxiliaries or nominal arguments.  

Talmy enumerates 35 different semantic components.  In addition to the six listed above, these include main  purpose, result, polarity, aspect,  personation, temporal and spatial setting, gender, valence,  etc. 
Talmy also isolates surface elements within a verb complex such as the root verb, inflections, appositions, subordinate clauses and satellites.
 He then examines which semantic elements are expressed by which surface elements. He finds that the relationship is mostly not one-to-one. A combination of semantic elements may be expressed by a single surface element, or a single semantic element by a combination of surface elements. In a similar manner, semantic elements of different types can be expressed by the same type of surface elements or by several different ones.

Talmy's work does not enumerate lexical entries for specific verbs, but provides   detailed discussion on  semantic facets of meanings of a verb.   The main thrust of Talmy's work  is to demonstrate that semantic elements and surface elements relate to each other in specific patterns, both typological and universal.  In work prior to Talmy's, most work has treated language's lexical elements as atomic givens, without involving semantic components that comprise them. These studies treated the properties that such whole forms can manifest, in particular, word order, grammatical relations and case roles. Talmy's  cross-linguistic  study determines semantic components' surface presence, site (their host constituent or grammatical relation) and combination within a site. In addition, Talmy's tracing of surface occurrence patterns extends beyond treating single semantic component at a time to treating a concurrent set of components. 

Lexical semantics must strive to represent at least some of the various semantic components that Talmy enumerates. In addition, it must incorporate ways of mapping from syntax to semantics or vice versa.  In a very simple system, a set of detailed rules may be able to enumerate the mappings from syntax to semantics and vice versa. In a complex modern system, it is necessary that a machine learning technique will automatically acquire the mappings. This usually requires a lot of labeled examples for a machine learning program to learn such mappings. We discuss some such as efforts later in the paper. 

\subsubsection{Jackendoff's Lexical Conceptual Structure}

\subsubsection{Generative Lexicon by Pustejovsky}
\label{section:GenerativeLexicon}

Consider the following illustrative examples.
\begin{itemize}
\item [a)] {\em Mary walked.}
\item [b)] {\em Mary walked to the store.}
\item [c)] {\em Mary walked for 30 minutes.}
\end{itemize}
Sentence {\em a)} describes  a process, which is an activity of of indefinite length, i.e., the sentence does not say how long the activity of walking took. Although {\em b)} does not give an explicit time duration for the walking event, it depicts an accomplishment and provides  a logical culmination to the duration of the  event of walking because the event is over when Mary reached the store. Sentence {\em c)} talks about  a bounded process in which, the event of walking terminating although it does not provide an explicit termination point, but provides a bound to the time extent of the activity in terms of a duration adverbial.  This example motivates the observation that the use of prepositional phrases or duration adverbials can change the (aspectual) classification of an event. To explain such phenomena better, it is beneficial to have more complex event structures or lexical analysis of event words. 
\cite{vanVoorst1988} hypothesizes that the direct object plays a role in delimitation of an event, i.e., whether it has a culmination or not. \cite{Pustejovsky1991a,Pustejovsky1991b} builds upon such observations and   hypothesizes that  it is necessary to know the how an event can be broken down into sub-events. He provides the following reasons for sub-eventual analysis.
\begin{itemize}
\item Sub-eventual analysis of predicates allows verbal decomposition leading to more complex lexical semantics. 
\item Scope of adverbial modification, for some adverbials, can be explained better using event sub-structures. 
\item Semantic arguments of items within a complex event structure can be mapped onto argument structures better. 
\end{itemize}
 Pustejovsky describes a {\em generative lexicon} in the sense that meanings are described in terms of a limited number of so-called generative devices or primitives by drawing upon Aristotle's {\em species of opposition} \cite{Lloyd1968}. For example, to express the meaning of the word {\em closed} as in {\em The door is closed}, {\em The door closed} or {\em John closed the door}, one needs the concept of opposition between {\em closed} and {\em not-closed}. This essential 
opposition in the meaning of a lexical item is described by Pustejovsky  in terms of what is called the {\em qualia structure} of the lexical item. Thus, there are three primary components to the event structure proposed by Pustejovsky.
\begin{itemize}
\item {\em Event type}: The event type of the lexical item is given in terms of the classification schemes discussed earlier. 
\item {\em Rules for event composition}: Since an event may be expressed by more than a single verb, the meanings of several lexical items may have to be composed to obtain a description. For example, how does PP attachment change the meaning of the central event in context?
\item {\em Mapping rules from event structure to argument structure}: Pustejovsky describes  a number of rules or principles for such mapping. These rules describe how how semantic participants are realized syntactically. 
\end{itemize}
Pustejovsky provides lexical meaning in terms of four separate structures.
\begin{itemize}
\item {\em Argument structure}: The behavior of a word as a function, with its arity. This provides the predicate argument structure for a word, which specifies how it maps to syntax.
\item {\em Event structure}: It identifies a specific event type for a word or a phrase, following \cite{Vendler1967}.
\item {\em Qualia structure}: It provides the essential attributes of an object that need to be expressed lexically.
\item {\em Inheritance structure}: It specifies how the word is globally related to other concepts in the lexicon. 
\end{itemize}

In summary,  Pustejovsky endows  complexity to lexical entries for verbs as well as non-verbs so that semantic weight does not fall on verbs alone in the lexicon and when composing the meaning of a sentence from its constituents. Pustejovsky's approach also reduces the number of lexical entries necessary for individual verbs because the lexical entries become more general. Pustejovosky focuses on creating more muscular compositional semantics rather than decomposing a verb's meaning into a specified number of primitives.

\subsection{Semantic Arguments and Syntactic Positions}

Frequently, specific semantic arguments of a verb (also called thematic arguments) appear in characteristic syntactic positions. This has led to theories or proposals regarding mapping between the two. These theories state that specific semantic arguments belong in specific syntactic positions and that there is 1-1 relationship between semantic argument and (initial\footnote{The syntactic or surface position of a constituent may change under certain circumstances, e.g., when one formulates a question from a canonical or declarative sentence.}) syntactic position. Such proposals or theories include the Universal Alignment Hypothesis \cite{Perlmutter1978} and Uniformity of Theta Assignment Hypothesis \cite{Baker1988}. These are supposed to be universal in that they applied across languages and across verbs. For example, agents appear in subject positions across languages and verbs. This mapping is thus universal. However, other mappings are not so universal. For example, the theme can appear in object, subject or indirect object position; and the experiencer can appear in subject or object position.

A theory that explains lexicon-to-syntax mapping also needs to explain the existence of argument alterations. In other words, it should explain the possibility that the same semantic role can appear in different syntactic positions for the same verb. Usually, linguists classify verbs into a number of semantic classes (different from the ones we talked about earlier) and for each class, a set of mapping relations and a set of argument alterations are specified \cite{Levin1993,Levin1995,Pinker1989}. However, other researchers claim that such semantic classification is difficult to obtain because semantically similar verbs may behave differently across languages \cite{Rosen1984}, a given verb in a language may have multiple syntactic realizations \cite{Rosen1984,Rosen1996}, and semantically similar verbs may allow several syntactic realizations \cite{Rosen1996}.

\section{Lexical Resources for Action or Event Representation}
The discussions on lexical representation of verbs so far have been based on efforts where a small number of examples were studied intently by linguists before making the various proposals. Starting the 1980s but more so in the 1990s, when computer scientists started to focus more on analysis of large text corpora, it became evident to some that the lexical analysis of pure linguists can be extended by knowledge gathered from such corpora. This led to development of  the Comlex lexicon \cite{Grishman1994}, WordNet \cite{Miller1995,Fellbaum2010}, VerbNet \cite{Schuler2005}, FrameNet \cite{Fillmore2001a,Fillmore2001b,Fillmore2003,Fillmore2006} and other resources. Some of these may have started without an automatic analysis of corpora, but soon corpora were used to refine and enhance the initial lexical resources.  Comlex was a substantial resource whose creators  spent a lot of effort in enumerating subcategorization features. WordNet is a large lexical resource or ontology, which encompasses words from all categories. WordNet includes verbs, but is not verb-specific. VerbNet, of course, is focussed on verbs alone. FrameNet is also focussed on verbs. Both VerbNet and FrameNet attempt to represent all verbs, not only those which are used to represent ``events". However, the term {\em event} itself is not clearly defined and most anything that is described by a verb can be considered an event in some context or another.

\subsection{Comlex and Nomlex Lexicons}
Comlex was created at New York University as a computational lexicon providing detailed syntactic information on approximately 38,000 words \cite{Grishman1994}.  Of course, not all of these were verbs or words that describe actions. The feature set Comlex provided were more detailed than commerically available dictionaries at the time such as the Oxford Advanced Learner's Dictionary (OALD)  \cite{Hornby1980} and Longman's Dictionary of Contemporary Englisch (LDOCE) \cite{Procter1981}.  The  initial word list was derived from OALD. The lexicon used a Lisp-like notation for dictionary entries. We see some sample entries for verbs in Comlex in Figure \ref{fig:ComlexExamples}. 

\begin{figure}
\begin{verbatim}
(verb :orth "abandon" :subc ((np-pp :pval ("to")) (np)))
(noun :oath "abandon" :features ((countable :pval ("with"))))
(verb :orth "abstain" :subc ((intrans) (pp :pval ("from")) (p-in-sc :pval ("from"))))
(verb :oath "accept" :subc ((np) (that-s) (np-as-np)))
(verb :orth ÒpromoteÓ :subc ( (np-pp :pval (ÒtoÓ ÒforÓ ÒintoÓ ÒfromÓ)) (np-pp-pp 
                     :pval (ÒforÓ ÒtoÓ ÒintoÓ ÒfromÓ)) (possing)(np)(np-as-np)(np-tobe)))
\end{verbatim}
\caption{Sample Comlex verb entries}
\label{fig:ComlexExamples}
\end{figure}

Comlex paid particular attention to providing detailed subcategorization or complement information for verbs, and nouns and adjectives that take complements. Comlex was influenced by prior  work on lexicon such as the Brandeis Verb Lexicon \cite{Grimshaw1981}, the ACUILEX project \cite{Sanfilippo1994}, the NYU Lingustic String Project \cite{Sager1981}, the OALD and the LDOCE, and it  incorporated distinctions made in these dictionaries. Comlex had 92 different subcategorization features for verbs. The features recorded differences in grammatical functions as well as constituent structure. In particular, Comlex captured four different types of control: subject control, object control, variable control and arbitrary control. It was also able to express the fact that a verb may have different control features for different complement structures, or different prepositions within the complement. Figure \ref{fig:ComlexComplements} shows a few complements used in Comlex. Here {\tt :cs} is the constituent structure, {\tt :gs} is the grammatical structure and {\tt :ex} are examples. The authors created a initial lexicon manually and then refined it using a variety of sources, both commercial and corpus-based. 

\begin{figure}
\begin{verbatim}
(vp-frame  s               :cs  ((s  2  :that-comp  optional)) 
                                    :gs  (:subject  1  :comp  2) 
                                    :ex  "they  thought  (that)  he  was  always  late") 

(vp-frame  to-inf-sc  :cs  ((vp  :2  :mood  to-infinitive  :subject  1))
                                    :features  (:control  subject) 
                                    :gs  (:subject  1  :comp  2) 
                                    :ex  "1 wanted  to  come.") 
 
(vp-frame  to-inf-rs   :cs  ((vp  2  :mood  to-infinitlve  :subject  1)) 
                                    :features  (:raising  subject) 
                                    :gs  (:subject  ()  :comp  2) 
                                    :ex  "they  seemed  to  wilt.")
\end{verbatim}
\caption{Sample Comlex Subcategorization Frames}
\label{fig:ComlexComplements}                                   
\end{figure}

The Nomlex dictionary of nominalizations was also developed at NYU \cite{Macleod1998,Meyers1998}. It enumerated allowed complements for nominalizations, and also related nominal complements of the corresponding verbs. A nominalization is the noun form of a verb. For example, the verb {\em promote} is nominalized as {\em nominalization}. Similarly, the nominalizations of the verb {\em appoint} are {\em appointment} and {\em appointee}. Nomlex entries are similar in syntax to Comlex entries. Each Nomlex entry has a {\tt :nom-type} feature which specifies four types of nominalizations: action ({\em appointment}, {\em destruction}) or state ({\em knowledge}), subject ({\em teacher}), object ({\em appointee}) and verb-part for those nominalizations that incorporate a verbal particle ({\em takeover}). Meyers et al. \cite{Meyers1998} presented a procedure what mapped syntactic and semantic information for an active clause containing a verb e.g., ({\em IBM appointed Alice Smith as vice president}) into a set of patterns for nominalization ({\em IBM's appointment of Alice Smith as vice president} or {\em Alice Smith's appointment as vice president}). 
The lexical entry for the verb {\em appoint} used in Comlex is given in Figure \ref{fig:ComlexExamples}. The lexical entry in Nomlex for the action nominalization {\em appointment} is given in Figure \ref{fig:NomlexExample}.

\begin{figure}
\begin{verbatim}
(nom :orth ÒpromotionÓ :verb ÒpromoteÓ
           :nom-type((verb-nom))
           :verb-subj ((n-n-mod) (det-poss))
           :verb-subc ((nom-np :object ((det-poss)(n-n-mod)(pp-of)))
                           (nom-np-as-np :object ((det-poss) (pp-of))) 
                           (nom-possing :nom-subc ((p-possing :pval (ÓofÓ)))) 
                           (nom-np-pp :object ((det-poss) (n-n-mod) (pp-of))
                                               :pval (ÓintoÓ ÓfromÓ ÓforÓ ÓtoÓ)) 
                           (nom-np-pp-pp:object ((det-poss) (n-n-mod) (pp-of))
                                                :pval (ÓforÓ ÓintoÓ ÓtoÓ) :pval2 (ÓfromÓ))))
\end{verbatim}
\caption{The lexical entry for the action nominalization {\em appointment} in Nomlex. The entry for the verb {\em appoint} is given in Figure \ref{fig:ComlexExamples}}.  
\label{fig:NomlexExample}
\end{figure}

\subsection{Levin's Verb Classes}
Levin's verb classes \cite{Levin1993} explicitly provide the syntax for each class, but do not provide semantic components. The classes are based on the ability or inability of a verb to occur in pairs of syntactic frames, with the assumption that syntactic frames reflect the underlying semantics. 
For example, {\em break} verbs and {\em cut} verbs are similar because they can all take part in transitive and middle constructions. However, only break verbs can occur in simple intransitive constructs. Similarly, {\em cut} verbs can occur in conative constructs and {\em break} verbs cannot. The explanation given is that {\em cut} describes a sequence of actions that result in the goal of separating an object into pieces. It is possible that one can perform the actions without achieving the result (e.g., {\em John cut at the loaf}). For {\em break}, the result is a changed state where the object becomes separated into pieces. If the result is not achieved, we cannot say that the action of breaking took place. The examples below are taken from \cite{Kipper2000b}.
\begin{itemize}
\item {\em Transitive Construction}: (a) {\em John broke the window.}, (b) {\em John cut the bread.}
\item {\em Middle Construction}: (a) {\em Glass breaks easily.}, (b) {\em This loaf cuts easily}.
\item {\em Intransitive Construction}: (a) {\em The window broke.}, (b) {*\em The bread cut.}
\item {\em Conative Construction}: (a) *{\em John broke at the window.}, (b) {\em John valiantly cut at the frozen loaf, but his knife was too dull to make a dent in it.}
\end{itemize}
Levin's original classes had some inconsistencies. For example, many verbs were listed in multiple classes, some of which had conflicting syntactic frames. \cite{Dang1998} refined the original classification to remove some of these problems to build a more fine-grained, syntactically and semantically coherent refined class called {\em intersective Levin classes}. Levin's classes also are focussed mostly on verbs taking noun (NP) and prepositional phrase (PP) complements, and are weak on coverage of ADJP, ADVP, sentential complement, etc. VerbNet is built using these classes. 

Organization of verbs into such classes capture generalizations about their properties. Such classes also help create better NLP systems. Many NLP systems benefit from using the mapping from surface realization of arguments to predicate-argument structure that  is available in such classes. These classes also capture abstractions (e.g., syntactic and semantic properties) and as a result, they are helpful in many operational contexts where the available corpora are small in size and thus, it is not possible to extract detailed lexical information about verbs from such small corpora. The predictive power of the classes can compensate for the lack of sufficient data. Lexical classes have been helpful in tasks such as subcategorization acquisition \cite{Dorr1997,Prescher2000,Korhonen2004}, automatic verb acquisition \cite{Swift2005}, semantic role labeling \cite{Swier2004}, and word sense disambiguation \cite{Dang2004}. 
***Add newer citations for application. Look at after 2004 proceedings of NAACL-HLT***

\subsection{WordNet}
The WordNet project \cite{Miller1995,Fellbaum2010} started in the mid-1980s at Princeton University and over time, has become the most widely used lexical resource in English, especially when one needs a lexical resource that can be used by a program. 
Wordnet was primarily designed as a semantic network and later modified to be a lexical database.  

WordNet groups words into synsets (synonym set)  and contains relations among these synsets. A synset contains all the word forms that can refer to a given concept or sense.  For each sense of each word, WordNet also provides a short, general definition called its {\em gloss} and example usages. 

As the name hints, the WordNet can be thought of as a large graph where the 
words and synsets are nodes. These nodes linked by edges that represent lexical and semantic-conceptual links, which we discuss briefly below. Individual words may also be linked with {\em antonym} links. Superclass-subclass relations link entire synsets. WordNet has entries for verbs, nouns, adjectives and adverbs. 

To get a better feel for what WordNet is like, let us look at the online version of WordNet\footnote{available at {\url{http://wordnetweb.princeton.edu}}} at Priceton University. 
When we search for the word {\em assault} in the online WordNet,  the results come in two parts: noun and verb, because {\em assault} can be either a verb or a noun. The results that show up for verb are given in Figure \ref{fig:WordNet1}. The verb senses of {\em assault} belongs to three synsets. In other words, it has three senses or can refer to three different concepts. Each sunset is composed of several verbs. The second of these synsets contains one sense of each of the verbs {\em assail}, {\em set on} and {\em attack}.  

 A verb may have four types of entries in WordNet: hypernyms, toponyms, entailment and coordinate terms. These terms are defined here. A  verb $Y$ is a {\em hypernym} of the verb $X$ if the activity $X$ is a (kind of) $Y$. For example, {\em to perceive} is an hypernym of {\em to listen}. 
A verb $Y$ is a troponym of the verb $X$ if the activity $Y$ is doing $X$ in some manner. For example, {\em to lisp} is a troponym of {\em to talk}. 
A verb $Y$ is entailed by $X$ if by doing $X$ one must be doing $Y$. For example, {\em to sleep} is entailed by {\em to snore}. 
{\em Coordinate terms} are  those verbs that share a common hypernym, e.g., {\em to lisp} and {\em to yell}. 
If we want to see the direct troponym of the second synset for the verb meaning of {\em assault}, we get what we see in Figure \ref{fig:WordNet2}.
 
\begin{figure}
\begin{verbatim}
S: (v) assail, assault, set on, attack (attack someone physically or 
    emotionally)
S: (v) rape, ravish, violate, assault, dishonor, dishonour, outrage (force 
   (someone) to have sex against their will)
S: (v) attack, round, assail, lash out, snipe, assault (attack in speech or 
   writing)
\end{verbatim}
\caption{Online WordNet search for the word {\em attack}. We show only the verb entries, with  gloss for each entry.} 
\label{fig:WordNet1}
\end{figure}

\begin{figure}
\begin{verbatim}
S: (v) bait (attack with dogs or set dogs upon)
S: (v) sic, set (urge to attack someone)
S: (v) bulldog (attack viciously and ferociously)
S: (v) rush (attack suddenly)
S: (v) blindside (attack or hit on or from the side where the 
     attacked person's view is obstructed)
S: (v) savage (attack brutally and fiercely)
S: (v) reassail (assail again)
S: (v) jump (make a sudden physical attack on)
S: (v) beset, set upon ((5) )
S: (v) rape, ravish, violate, assault, dishonor, dishonour, outrage 
     (force (someone) to have sex against their will)
S: (v) desecrate, profane, outrage, violate (violate the sacred 
     character of a place or language)
S: (v) molest (harass or assault sexually; make indecent advances to)
\end{verbatim}
\caption{The direct toponym set for the second sunset for the verb {\em assault} seen in Figure \label{fig:WordNet2}}. 
\end{figure}

WordNet has been used in many applications. However,  it is  most commonly used as a computational lexicon or ``ontology" of English (or, another language)  for word sense disambiguation, a task that assigns the most appropriate senses (i.e. synsets) to words in specific contexts. 
Although WordNet is large and detailed, WordNet does not have information required by NLP applications such as predicate-argument structure. Although WordNet contains a sufficiently wide range of common words, it does not cover special domain vocabulary. It is general in nature, and therefore difficult to use if specialized vocabulary is needed. Also, WordNet senses are sometimes overly  fine-grained even for human beings and as a results, some researcher argue that it cannot achieve very high performance in the tasks where it is applied.  
Although WordNet is the  most widely used online lexical database in NLP applications, is also limited in its coverage of verbs. 

The English WordNet  currently contains approximately  117,659  synsets, each sunset corresponding to a sense of a word.
It has 11,529 verbs that belong to 13,767 synsets. It also contains 117,798 nouns that belong to 82,115 synsets. WordNets have been developed or are being developed in a large number of languages such as  Catalan, French, Spanish, Japanese, Chinese, Danish, Korean and Russian.  Notable collaborative efforts include Euro Wordnet \cite{Vossen2004,Vossen1998a,Vossen1998b}, Asian Wordnet  \cite{Charoenporn2008,Robkop2010,Sornlertlamvanich2009} and Indo WordNet \cite{Sinha2006} projects. The Indo WordNet focuses on 18 major languages of India. For example, as of June 2012\footnote{\url{http://en.wikipedia.org/wiki/IndoWordNet}} there are 15,000 synsets in the Assamese WordNet, 24,000 in  Bengali, 16,000 in  Bodo, 27,000 in  Gujarati, and 31,500 in  Oriya. WordNets in most other languages are not as sophisticated as the one in English.

\subsection{FrameNet}
FrameNet  \cite{Baker1998,Fillmore2003,Ruppenhofer2006} is another substantial publicly available lexical resource that has come into existence independently. It is based on the theory of frame semantics \cite{Fillmore1976,Petruck1996,Fillmore2001a,Fillmore2001b,Fillmore2006} where a {\em frame} corresponds to a stereo-typical scenario involving an {\em interaction} and {\em participants}, where participants play some kind of roles. The idea is that the meanings of most words are best understood in context. FrameNet proposes a small context, called a {\em semantic frame}, a description of a type of event, relation or entity and the participants in it. 
 A frame has a name and this name is used to identify a semantic relation that groups together the semantic roles.

 Although frames mostly correspond to verbs, there are frames that can be identified by nouns and adjectives. FrameNet also has a large number of annotated sentences. Each annotated sentence exemplifies a possible syntactic realization of the semantic role associated with a frame for a given target word. FrameNet extracts syntactic features and corresponding  semantic roles from all annotated sentences in the FrameNet corpus, it builds a large set of rules that encode possible syntactic realizations of semantic frames. 

FrameNet aims to document the range of semantic and syntactic combinatory possibilities--- {\em valences}--of each word in each of its senses, through computer-assisted annotation of example sentences and automatic tabulation  of the annotation results. The FrameNet lexical database, currently contains more than 10,000 lexical units (defined below), more than 6,000 of which are fully annotated, in nearly 800 hierarchically-related semantic frames, exemplified in more than 170,000 annotated sentences. See the FrameNet website\footnote{\url{https://framenet.icsi.berkeley.edu/fndrupal/home}} for the latest statistics. 
FrameNet has been used as a semantic role labeling, used in applications such as information extraction, machine translation, event recognition, sentiment analysis, etc., like the other publicly available lexical resources. 

An example of a frame is {\em Attack}. This frame has several {\em frame elements}. The {\em core} frame elements are {\em assailant} and {\em victim}. There are a large number of non-core frame elements. These include {\em Circumstances}, {\em Containing\_event}, {\em Direction}, {\em Duration}, {\em Explanation}, {\em Frequency}, {\em Manner}, {\em Means}, {\em Place}, {\em Purpose}, {\em Result}, {\em Time}, {\em Weapon}, etc. For each of these frame elements there can be seen in one or more annotated sentences. Here is an example annotated sentence. 

\begin{multline}
[_{Assailant} \; The \; gang] \; ASSAULTED  [_{Victim} \; him] \\
[_{Time} \; during \; the \;drive \; to \; Rickmansworth] [_{Place} \; in \;Hertfordshire] ...
\end{multline}

The frame {\em Attack} is associated with a large number of associated units. These include verbs and nouns. Example verbs are {\em ambush},  {\em assail}, {\em assault}, {\em attack}, {\em bomb}, {\em bombard}, {\em charge}, {\em hit}, {\em infiltrate}, {\em invade}, {\em raid}, {\em storm} and {\em strike}. Examples of nouns are {\em airstrike}, {\em ambush}, {\em assailant}, {\em assault}, {\em attack}, etc. The frame {\em Attack} inherits from a frame called {\em Intentionally\_affect}. It is inherited by frames {\em Besieging}, {\em Counterattack}, {\em Invading} and {\em Suicide\_attack}. 

FrameNet annotates each frame element (or its representation, actually) in at least three layers: a frame element name (e.g., {\em Food}), a grammatical function (e.g., {\em Object)} and a phrase type (e.g., {\em NP}). Only the frame elements are shown in the Web-based interface to reduce visual clutter, although all three are available in the XML downloads. FrameNet has defined more than 1000 semantic frames. These frames are linked together using frame relations which relate more general frames to specific ones. This allows for reasoning about events and intentional actions. 

Because frames are semantic, they are often similar across languages. For example, frames about buying and selling involve frame elements {\em Buyer}, {\em Seller}, {\em Goods} and {\em Money} in every language. FrameNets have been developed for languages such as Portuguese, German, Spanish, Chinese, Swedish and Japanese. 

At the current time, there are 1159 frames in FrameNet. There are approximately 9.6 frame elements per frame. There are 12595 lexical units of which 5135 are nouns, 4816 are verbs, 2268 are adjectives. There are 12.1 lexical units per frame. 

There have been some attempts at extending the coverage of FrameNet. One such effort is by \cite{Rastogi2014} who use  a new broad-coverage lexical-semantic resource called PPDB to add lemmas as pontential triggers for a frame and to automatically rewrite existing example sentences with these new triggers. PPDB, The Paraphrase Database, is a lexical, phrasal and syntactic paraphrase database \cite{Ganitkevitch2013}. They use PPDB's lexical rules along with a 5-gram Kneser-Ney  smoothed language model trained using KenLM \cite{Heafield2013} on the raw English sequence of the Annotated Gigaword corpus \cite{Napoles2012}. 

\subsection{PropBank}
PropBank \cite{Kingsbury2002,Kingsbury2003,Palmer2005} is an annotated corpus of verb propositions and their arguments. PropBank does not annotate events or states of affairs described using nouns. PropBank-style annotations usually are closer to the syntactic level, whereas FrameNet-style annotations are more semantically motivated although, as discussed earlier, FrameNet provides layers of annotations including syntactic parses. 
PropBank annotates one million words of the Wall Street Journal portion of the Penn Treebank \cite{Marcus1994} with predicate-argument structure for verbs using semantic role labels for each verb argument. 

Although the same tags are used across all verbs (viz., Arg0, Arg1, $\cdots$, Arg5), these tags have verb-specific meaning. FrameNet requires that the use of a given argument label is consistent across different uses of a specific verb, including its syntactic alternations. Thus,  Arg1 (italicized) in ``John broke {\em the window} broke" is the same window that is annotated as the Arg1 in ``{\em The window} broke" even though it is the syntactic subject in one case and the syntactic object in another. FrameNet does not guarantee that an argument label is used consistently across different verbs. For example, Arg2 is used as label to designate the {\em destination} of the verb ``bring", but the {\em extent} of the verb ``rise". Generally, the arguments are simply listed in the order of their prominence for each verb. However, PropBank tries to use Arg0 as the consistent label for the ``prototypical agent" and Arg1 for the ``prototypical patient" as discussed in \cite{Dowty1991}. 

PropBank divides words into lexemes using a very coarse-grained sense disambiguation scheme. Two senses are considered distinct only if their argument labels are different. In PropBank each word sense is called a {\em frameset}. PropBank's model of predicate-argument structure differs from dependency parsing. In dependency parsing, each phrase can be dependent only on one other phrase. But, in PropBank, a single phrase can be arguments to several predicates.  PropBank provides a lexicon which divides each word into coarse-grained senses or framesets, and provides examples usages in a variety of contexts.  For example, the {\em to make an attack, criticize strongly} sense of the predicate lemma (or, verb) {\em attack} is given in Table \ref{table:PropBank} along with an example.

\begin{table}
\begin{tabular}{| l | l | l | } \hline
PropBank Role & Meaning of PropBank Role & VerbNet (Theta) Role \\ \hline
Arg0 & attacker & Agent \\
Arg1 & entity attacked& Theme \\
Arg2 & attribute & Predicate \\ \hline \hline
\multicolumn{3}{|l|}{Example: {\em Mr. Baldwin is attacking the greater problem: lack of ringers.}} \\ \hline
PropBank Role & \multicolumn{2}{|l|}{Value}\\ \hline 
Arg0 & \multicolumn{2}{|l|}{Mr. Baldwin}\\ \hline
Arg1 & \multicolumn{2}{|l|}{the greater problem: lack of ringers}\\ \hline
\end{tabular}
\caption{PropBank entry for {\em attack.01}, a sense of the verb {\em attack}}
\label{table:PropBank}
\end{table}

PropBank tries to keep rolesets consistent across related verbs. Thus, for example, the {\em buy} roleset is similar to the {\em purchase} and {\em sell} rolesets. See Table \ref{table:PropBankBuySell}, taken from \cite{Kingsbury2002}.

\begin{table}
\begin{center}
\begin{tabular}{| l | l | l|}\hline
{\bf purchase} & {\bf buy} & {\bf sell}\\ \hline
Arg0: buyer & Arg0: buyer & Arg0: seller \\
Arg1: thing bought & Arg1: thing bought & Arg1: thing sold\\
Arg2: seller & Arg2: seller & Arg2: buyer\\
Arg3: prince paid & Arg3: price paid & Arg3: price paid\\
Arg4: benefactive & Arg4: benefactive & Arg4: benefactive\\ \hline
\end{tabular}
\end{center}
\caption{Roles for the verbs {\em purchase}, {\em buy} and {\em sell} in PropBank}
\label{table:PropBankBuySell}
\end{table}
One can clearly see that it may be possible to merge such similar framesets together to obtain something similar to the verb roles in FrameNet's {\em Commerce} frameset. 

Although similar, PropBank differs from FrameNet we have discussed earlier in several ways. PropBank is a resource focussed on verbs whereas FrameNet is focussed on frame semantics that generalizes descriptions across similar verbs as well as nouns and other words (e.g., adjectives) as discussed earlier. PropBank was created with the idea of serving as training data to be used with machine learning algorithms for the task of semantic role labeling. It requires all arguments to a verb to be syntactic constituents in nature. In addition, PropBank differentiates among senses of a verb if the senses take different sets of arguments. There is a claim that due to such differences, semantic role labeling is easier using a corpus annotated with PropBank type annotation compared to FrameNet type annotation. 

\subsection{VerbNet}
VerbNet \cite{Kipper2000a,Kipper2000b,KipperSchuler2005} attempts to provide a definitive resource for lexical entries for English verbs. It is compatible with WordNet, but has lexical entries with explicit syntactic and semantic information about verbs, using Levin's verb classes \cite{Levin1993}. It uses verb classes to capture generalizations and for efficient encoding of the lexicon. Its syntactic frames for verb classes are represented  using a fine-grained variation of Lexicalized Tree Adjoining Grammers \cite{Joshi1985,Schabes1990,Dang1998} augmented with semantic predicates, allowing for creating compositional meanings for more complex constituents such as phrases and clauses. VerbNet provides traditional semantic information such as thematic roles and semantic predicates, with syntactic frames and selectional restrictions. it also allows for extension of verb meaning through adjunction of particular syntactic phrases. 

A verb entry corresponds to a set of classes, corresponding to the different senses of the verb. For each verb sense, there is a verb class as well as specific selectional restrictions and semantic characteristics that may not be captured by class membership. VerbNet also contains references to WordNet synsets. Verb classes capture generalizations about verb behavior. Each verb class lists the thematic roles that the predicate-argument structure of its members allows, and provides descriptions of the syntactic frames corresponding to allowed constructs, with selectional restrictions given for each argument in each frame. Verb classes are hierarchically organized. It required some manual restructuring of Levin's classes. Each event $E$ is decomposed into a three-part structure according to \cite{Moens1987,Moens1988}. VernNet uses a time function for each predicate specifying whether the predicate is true during the preparatory, culmination or consequent/result stage of an event. This structure allows VerbNet to express the semantics of classes of verbs like {\em Change of State} verbs. For example, in the case of the verb {\em break}, it is important to distinguish between the state of the object before the end of the action and the new state that results afterwards. 

Table \ref{table:VerbNetExample} is an example of a simplified VerbNet entry from its website\footnote{\url{http://verbs.colorado.edu/~mpalmer/projects/verbnet.html}}. The original VerbNet was extended using extensions proposed by \cite{Korhonen2004}. This resulted in the addition of a large number of new classes, and also a much more comprehensive coverage of English verbs. Table \ref{table:VerbNetStatistics} provides statistics of VerbNet's coverage in its initial version, VerbNet as described in \cite{Kipper2000a,Kipper2000b,Kipper2008}, and its current version as in its official Website. 

\begin{table}
\begin{tabular}{| p{0.5in} | p{.5in} | p{.5in} | p{2.75in} |}\hline
\multicolumn{4}{|c|}{Class Hit-18.1} \\ \hline
\multicolumn{4}{|l|}{Roles and Restrictions: Agent[+int\_control]  Patient[+concrete] Instrument[+concrete]} \\ \hline
\multicolumn{4}{|l|}{Members: bang, bash, hit, kick, $\cdots$}\\ \hline
\multicolumn{4}{|l|}{Frames: }\\ \hline
Name & Example & Syntax & Semantics \\ \hline
Basic Transitive & Paula hit the ball & Agent V Patient & cause(Agent, E) manner(during(E), directedmotion, Agent)\newline 
!contact(during(E), Agent, Patient) manner(end(E),forceful, Agent) \newline
contact(end(E), Agent, Patient) \\ \hline
\end{tabular}
\caption{Simplified VerbNet entry for Hit-18.1 class. Every class is numbered.}
\label{table:VerbNetExample}
\end{table}

\begin{table}
\begin{center}
\begin{tabular}{|  l  |  l  | p{.75in}  |  l | } \hline
 & VerbNet1.0   & VerbNet  \cite{Kipper2000a} & VerbNet now \\ \hline
 First-level classes & 191 & 237 & 274 \\ \hline
 Thematic roles & 21 & 23 & 23\\ \hline
 Semantic predicates & 64 & 94 & 94\\ \hline
 Selectional restrictions (semantic) & 36 & 36 &?\\ \hline
 Syntactic restrictions & 3 & 55 &55\\ \hline
 Lemmas & 3007 & 3175 &3769 \\ \hline
 Verb senses & 4173 & 4526 &5257 \\ \hline
\end{tabular}
\end{center}
\caption{VerbNet statistics}
\label{table:VerbNetStatistics}
\end{table}

The absence of any lexicon or resource that provides for accurate and comprehensive predicate-argument structure (or semantic role labels) for English verbs has been long considered a critical element that was needed to produce robust natural language processors. This was shown clearly by \cite{Han2000} who evaluated an English-Korean machine translation system. The authors showed that among several factors impacting on the low quality of translations, one that was most influential was the inability to  predicate-argument structure. Even with a grammatical parse of the source sentence ad complete vocabulary coverage, the translation was frequently bad. This is because, the authors found, that although the parser recognized the constituents that are verb arguments, it was unable to precisely assign the arguments to appropriate positions. This led to garbled translations. Simply preserving the proper argument position labels and not changing other things, resulted in substantial improvement in acceptable translations. When using one parser, the improvement was 50\%; with a second parser, the improvement was dramatic 300\%. Thus, the purpose in developing lexical resources such as FrameNet and PropBank, PropBank especially so, is to provide for training data annotated with predicate-argument positions with labels. Such data can be used with machine learning techniques. 

\subsection{Combining FrameNet, VerbNet and WordNet}
There have been attempts to integrate lexical resources to obtain more robust resources with wider coverage. We discuss one such effort here. \cite{Shi2005} integrate FrameNet, VerbNet and WordNet discussed earlier into a single and richer resource with the goal of enabling  robust semantic parsing. The reason for building connections among the three lexical resources is that similar syntactic patterns often introduce different semantic interpretations and similar meanings can be realized in many different ways.  The improved resource provides three  enhancements: (1) It extends the coverage of FrameNet, (2) It augments VerbNet's lexicon with frame semantics, and (3) It implements selectional restrictions using WordNet semantic classes.  
They use knowledge about words and concepts from WordNet,  information about different situations from FrameNet, and verb lexicon with selectional restrictions from VerbNet. They extract syntactic features and corresponding semantic roles from all annotated sentences in FrameNet to build a large set of rules that encode the possible syntactic realization of semantic frames. They identify the VerbNet verb class that corresponds to a FrameNet frame and this allows them to parse sentences that include verbs not covered by FrameNet. This they do by exploiting a transitivity relation via VerbNet classes: verbs that belong to the same Levin classes are likely to share the same FrameNet frame, and their frame semantics can be analyzed even if not explicitly defined in FrameNet. They use information from WordNet in several stages in the parsing process. The argument constraints encoded in VerbNet (e.g., $+animate, +concrete$) are mapped to WordNet semantic classes, to provide selectional restrictions for better frame selection and role labeling in a semantic parser.  In addition, the mapping between WordNet verb entries and FrameNet lexical units allows them to extend the parser coverage, by assigning common frames to verbs that are related in meaning according to the WordNet semantic hierarchies. The authors found that their mapping algorithms produced 81.25\% correct assignment of VerbNet entries with a correct FrameNet frame. They also were able to map 78.22\% VerbNet predicate-argument  structures with some syntactic features and selectional restrictions to the corresponding FrameNet semantic roles.

\subsection{OntoNotes and Other Large-scale Annotated Corpora}
The OntoNotes project \cite{Hovy2006,Sameer2007,Weischedel2010} has created an infrastructure for much richer domain independent representation of shallow meaning for use in natural language processing tasks, including event detection and extraction,  in English, Chinese and Arabic. OntoNotes annotates documents at several layers: syntax, propositions, word senses including nominalizations and eventive noun senses, named entities, ontology linking and co-reference. It has been designed to be a well-annotated large-scale corpus from which machine learning programs can learn many different aspects of meaning felicitously. 

OntoNotes uses Penn TreeBank parses \cite{Marcus1993}, PropBank propositional structures \cite{Kingsbury2002,Kingsbury2003,Palmer2005} on top of Penn Treebank, and uses the Omega ontology \cite{Philpot2005}  for word sense disambiguation. As we know, the Penn Treebank is annotated with information from which one can extract predicate-argument structures. The developers of OntoNotes  use a parser that recovers these annotations \cite{Gabbard2006}.  The Penn Treebank also has markers for ``empty" categories that represent displaced constituents. Thus, to create OntoNotes, its developers use another parser \cite{Collins1999,Collins2003} to extract function words. They also use a maximum entropy learner and voted preceptons to recover empty categories. PropBank, as we know, annotates the one-million word Wall Street Journal part of  the Penn Treebank with semantic argument structures for verbs. As we have noted earlier, the creators of OntoNote and others have discovered that WordNet's very fine grained sense distinctions make inter-annotator agreement or good tagging performance difficult. To achieve better performance, OntoNotes uses a method \cite{Palmer2004,Palmer2007}  for sense inventory creation and annotation that includes links between grouped word senses and the Omega ontology \cite{Philpot2005}. OntoNotes represents sense distinctions in a hierarchical structure, like a decision tree, where coarse-grained distinctions are made at the root and increasingly fine-grained restrictions until reaching WordNet senses at the leaves. Sets of senses under specific nodes of the tree are grouped together into single entries, along with syntactic and semantic criteria for their groupings; these are presented to annotators for improved annotation agreement, obtaining up to 90\% inter-annotator agreement. 
OntoNote follows a similar method for annotation of nouns.

To allow access to additional information such as subsumption, property inheritance, predicate frames from other sources, links to instances and so on, OntoNotes also links to an ontology. This requires decomposing the hierarchical structure of OntoNotes into subtrees which then can be inserted at the appropriate conceptual node in the ontology. OntoNotes represents its terms in the  Omega ontology \cite{Philpot2005}. Omega\footnote{\url{http://omega.isi.edu}} has been  assembled by merging a variety of sources such as WordNet, Mikrokosmos \cite{Mahesh1995}, and a few upper ontologies such as DOLCE \cite{Gangemi2002}, SUMO \cite{Niles2001}, and  Penman Upper Model \cite{Hovy2003}. OntoNote also includes and cross-references verb frames from PropBank, FrameNet, WordNet and Lexical Conceptual Structures \cite{Habash2002}.  
OntoNotes also has coreferences. It connects coreferring instances of specific referring expressions, primarily NPs that introduce or access a discourse entity. 

For the purpose of our paper, it is important to know that  OntoNotes tries to annotate nouns that carry predicate structure, e.g., those whose structure is derived from their verbal form. In particular, OntoNotes annotates nominalization and eventive senses of nouns. OntoNotes applies two strict criteria for identifying a sense of a noun as a  {\em nominalization} \cite{Weischedel2010}.
\begin{itemize}
\item The noun must relate transparently to a verb, and typically display a nominalizing morpheme such as {\em -ment (govern/government)}, {\em -ion (contribute/contribution)}, though it allows some zero-derived nouns such as $kill$, the noun derived from $kill$, the verb. 
\item The noun must be able to be used in a clausal noun phrase, with its core verbal arguments related by semantically empty or very ``light" licensers, such as genitive markers (as in {\em The Roman's destruction of the city..} or with the verb's usual particle or prepositional satellites as in {\em John's longing for fame and fortune...}
\end{itemize}

Just like nominalization senses, OntoNotes has strict definition of {\em eventive} senses. They have two definitional criteria (1) and (2),  and a diagnostic test (3), for determining if a noun sense is eventive. 
\begin{itemize}
\item [(1)] Activity causing a change of state: A noun sense is eventive when it refers to a single unbroken activity or process, occurring during a specific time period, that effects a change in the world of discourse. 
\item [(2)] Reference to activity proper: The noun must refer to the actual activity or process, not merely to the result of the activity or the process. 
\item [(3)] The noun patterns with eventive predicates in the ``have" test: \cite{Belvin1993} describes the following heuristic lexico-syntactic diagnostic test to apply to many nouns. The test has four parts to it as discussed briefly below.
\begin{itemize}
	\item Create a natural sounding sentence using the construction {\em X had {\textless}NP\textgreater} where {\textless}NP\textgreater  \;  is a 	noun phrase headed by the noun in question, e.g., {\em John had a party}.
	\item Check if the sentence can be used in present progressive as in {\em John is having a party.}  If the  sentence is felicitous, it adds to the noun being inventive. If it sounds odd, it adds to the evidence that the noun is stative. 
	\item Check if the sentence can be used in a pseudo-cleft construction such as {\em What John did was have a party.} If it is felicitous, the noun is more likely to be eventive. If not, it is more likely to be stative.
	\item Check if the sentence suggests  iterative or habitual action using the simple present such as  {\em John has a party every Friday}. If so, it adds evidence that the noun is eventive. If the sentence suggests that the situation is taking place at that very moment that it is uttered, it adds evidence that the noun is stative as in {\em John has a cold}. 
\end{itemize}
\end{itemize}

In addition to OntoNotes, there have been other efforts at obtaining large-scale annotated corpora such at the GLARF project  \cite{Meyers2001} that tries to capture information from various Treebanks and superimpose a predicate argument structure. The Unified Linguistic Annotation (ULA) project \cite{Pustejovsky2005} is a collaborative effort that aims to merge PropBank, NomBank, the Penn Discourse Treebank \cite{Prasad2008} and TimeBank \cite{Pustejovsky2003b} with co-reference information. 

\section{Extracting Events from Textual Documents}
\label{section:eventExtraction}
Different models of events have been used 
in computational linguistics work geared toward information extraction. 
\begin{itemize}
\item A type of model that has been carefully developed over many years,  treats 
 an event as a word that points to a node in a network of predominantly temporal relations. An example is the so-called TimeML event that is found in documents that are annotated using the TimeML guidelines \cite{Pustejovsky2003,Sauri2005a,Sauri2005}. In a TimeML annotated corpus, every event is annotated. Thus, when working with the TimeML model, an extraction program attempts to extract every event. 

\item A second  type of event model enumerates a few types of events to be extracted (where the types selected in a somewhat ad-hoc manner) with  an event being described or pointed to by one or more words, along with additional associated words or phrases that  specify arguments of  the event. 
An example is  the ACE model of events \cite{}, where an event is a complex structure with  arguments which themselves may be complex structures. 
 Event extraction in the context of MUC-7 (are there other related MUCs?) or ACE,  requires one to extract only a limited number of event types, e.g., movement event type, conflict event type, justice event type (see Section... below). The structure for these event types is provided by the contest organizers. 
 
 \item Authors have used very specialized definitions of events when working with biomedical text, with details of a few specialized types of biomedical events, along with the participants. 
 
 \item When detecting events in informal very short text such as microblogs of Facebook posts, researchers have used definitions that focus on extracting events from many short documents over which an event may be described. 
\end{itemize}
We discuss TimeML events next followed by  events. We discuss biomedical event extraction in Section \ref{}, and extraction of events from Twitter in Section \ref{}. 
 
 \subsection{TimeML Events}
 TimeML is a rich specification language for event and temporal expressions in natural language text. 
 In the TimeML \cite{Pustejovsky2003,Sauri2005a} annotation scheme, an event is a general term for situations that {\em happen} or {\em occur}. Events can be punctual or momentary, or last for a period of time. Events in TimeML format may also include predicates describing {\em states} or {\em circumstances} in which something holds true. Only those states that participate in an opposition  structure, as discussed in Subsection
  \ref{section:GenerativeLexicon}, are annotated.  In general, an event can be expressed in terms of verbs, nominalizations, adjectives, predicative clauses, or prepositional phrases. TimeML allows an event, annotated with the {\tt EVENT} tag, to be  one of seven types: occurrence, state, report, i-action, i-state, aspectual and perception. The first five are special cases. The last two, {\em Occurrence} and {\em State} are used for general cases that do not fall in the special ones. 
 
 \begin{itemize}
 
 \item Reporting: A reporting event describes an action declaring
something, narrating an event, informing about a situation, and so on.
Some verbs which express this kind of event are {\em say}, {\em report}, {\em tell}, {\em explain}, and {\em state}. An example sentence with the verb {\em say} is {\em Punongbayan {\bf said} that the 4,795-foot-high volcano was spewing gases up to 1,800 degrees.}

 \item I-Action: {\em I} stands for {\em intensional}\footnote{According to the English Wikipedia: In logic and mathematics, an intensional definition gives the meaning of a term by specifying all the properties required to come to that definition, that is, the necessary and sufficient conditions for belonging to the set being defined.}. 
 According to the TimeML annotation guidelines, an i-action is a dynamic event that takes  an event-denoting argument, which must be explicitly present in the text. Examples of verbs that are used to express i-actions include {\em attempt}, {\em try}, {\em promise} and {\em offer}. An example sentence with the verb {\em try}  is {\em Companies such as Microsoft or a combined worldcom MCI are {\bf trying} to monopolize Internet access.}
 
 \item I-State: I-State stands for {\em intensional state}. Like an  I-Action, an I-State event takes an
argument that expresses an event. Unlike an I-Action, the I-State class is
used for events which are states. An example sentence that uses the verb {\em believe} is {\em We {\bf believe} 
 that his words cannot distract the world from the facts of Iraqi aggression}. Other verbs used to express i-states include {\em intend}, {\em want}, and {\em think}. 

 \item Aspectual: An aspectual predicate takes an event as an argument, and points to a part of the temporal structure of the event. Such a part may be the beginning, the middle or the end of an event. Verbs such as {\em begin}, {\em finish} and {\em continue} are such aspectual predicates. An example sentence with the verb {\em begin} is {\em All non-essential personnel should {\bf begin} evacuating the sprawling base.} 

 \item Perception: 
 This class includes events involving the physical perception of another event.
Such events are typically expressed by verbs such as {\em  see}, {\em watch}, {\em glimpse}, {\em  hear},
{\em listen}, and {\em overhear}. An example sentence with the verb {\em see} is {\em Witnesses tell Birmingham police they {\bf saw} a man running.}

 \item Occurrence:  An occurrence is a general event that occurs or happens in the world.  An example of an occurrence is given in the following sentence: {\em The Defense Ministry said 16 planes have {\bf landed} so far with protective equipment against
biological and chemical warfare.} The occurrence has been highlighted in bold. 
 
 \item State: A state describes circumstances in which something obtains or holds true. An example sentence that shows two states is {\em  It is the US economic and political {\bf embargo} which has {\bf kept} Cuba in a box.}

 \end{itemize}
 
 TimeML allows one to mark up temporal expressions using the {\tt TIMEX3} tag. Temporal expressions are of three types: (a) Fully specified temporal expressions such as {\em June 11, 2013}, (b) Underspecified temporal expressions such as {\em Monday}, (c) Durations such as {\em three days}.  TimeML uses the {\tt SIGNAL} tag to annotate sections of text, usually function words, that indicate how temporal objects are related to each other. The material marked by SIGNAL may contain different types of linguistic elements: indicators of temporal relations such as prepositions such as {\em on} and {\em during}, other temporal connectives such as {\em when}, etc. The {\tt TIMEX3} and {\tt SIGNAL} tags were introduced by \cite{Setzer2000,Setzer2001}. 
 
 A major innovation of TimeML is the {\tt LINK} tags that encode relations between temporal elements of a document and also help establish ordering between the events in a document. There are three types of links: {\tt TLINK} showing temporal relationships between events, or between an event and a time; {\tt SLINK} or a subordination link to show context that introduces relations between two events, or an event and a signal; {\tt ALINK} or an aspectual link to show relationship between an aspectual event and its argument event. {\em TLINK} allows for 13 temporal relations introduced by \cite{Allen1983,Allen1984}. {\tt SLINK} is used to express contexts such as use of modal verbs, negatives, positive and negative evidential relations, factives which require the event argument to be true, and counterfactives which require the event argument to be false. {\tt ALINK} expresses initiation, culmination, termination or continuation relationships between an event and its argument event. Finally, TimeML is able to express three types of causal relations: an event causing an event, an entity causing an event, 
and the special situation where the use of the discourse marker {\em and} as a signal to introduce a {\tt TLINK} indicating that one event happened before another as in {\em He kicked the ball {\bf and} it rose into the air}. 

The creators of TimeML have spent significant efforts to develop a fairly large corpus annotated with TimeML tags. This corpus is called the TIMEBANK corpus  \cite{Pustejovsky2003b} and has 300 annotated articles. This corpus has been used to learn to extract events and temporal relations among events. 

\subsection{ACE Events}
\label{subsection:ACEEvents}
In the ACE model, only ``interesting'' events are annotated in corpora and thus extracted by a trained program. 
ACE annotators specify the event types they want to be extracted.
For example, in one information extraction contest, an ACE 2005 event\footnote{\url{http://projects.ldc.upenn.edu/ace/docs/English-Events-Guidelines_v5.4.3.pdf}} was of 8 types,  each with one has one or more sub-types. The types are given below. ***Maybe, give some examples***
\begin{itemize}
\item {\em Life}: Be-born, marry, divorce, injure and die
\item {\em Movement}: Transport
\item {\em Transaction}: Transfer-ownership, Transfer money
\item {\em Business}: Start-organization, Merge-organization, Declare-bankruptcy
\item {\em Contact}: Meet, Phone-write
\item {\em Conflict}: Attack, demonstrate
\item {\em Personnel}: Start position, End position, Nominate, Elect,  and 
\item {\em Justice}: Arrest-Jail, Release-Parole, Trial-Hearing, Charge-Indict, Sue, Convict, Sentence, Fine, Execute, Extradite, Acquit, Appeal, Pardon. 
\end{itemize}
Each event also has four  categorial attributes. The attributes and their values are given below.
\begin{itemize}
\item {\em Modality}: Asserted and Other where Other includes, but is not limited to: Believed events; Hypothetical events; Commanded and requested events; Threatened, Proposed and Discussed events; and Promised events.
\item {\em Polarity}: Positive and Negative.
\item {\em Genericity}: Specific, Generic
\item {\em Tense}: Past, Present, Future and Unspecified. 
\end{itemize}
ACE events have arguments. Each event type has a set of possible argument roles, which may be filled by entities, time expressions or other values. Each event type has a set of possible argument roles. There are a total of 35 role types although no single event can have all 35 roles. A complete description of which roles go with which event type can be found in the annotation guidelines for ACE 2005 events\footnote{\url{http://projects.ldc.upenn.edu/ace/docs/English-Events-Guidelines_v5.4.3.pdf}}. 
In an ACE event, time is noted if when explicitly given. 

Others have defined events or event profiles themselves to suit their purpose. For example,  Cybulska and Vossen \cite{Cybulska2010,Cybulska2011} describe an historical information extraction system where they extract event and participant information from Dutch historical archives. They extract information using what they call profiles. For example, they have developed 402 profiles for event extraction although they use only 22 of them in the reported system. For extraction of participants, they use 314 profiles. They also 43 temporal profiles and 23 location profiles to extract temporal and locational information. Profiles are created using semantic and syntactic information as well as information gleaned from Wordnet \cite{Miller1995}.

\subsubsection{Additional Annotation Schemes}
\paragraph{ERE Annotations}
The ACE annotation scheme, discussed earlier,  was developed by NIST in 1999, and the ERE (Entities, Relations and Events) scheme was defined as a simpler version of ACE \cite{Aguilar2014}. One of ERE's goals is also to make annotating easier and annotations more consistent across annotators. ERE attempts to achieve these goals by removing the most problematic annotations in ACE and consolidating others. We will discuss the three types annotations now: Entities, Relations and Events. 

For example, consider Entities. ACE and ERE  both have Person, Organization, Geo-Political Entity and Location as types of entities. ACE has two additional types, Weapon and Vehicle, which ERE does not have. ERE doesn't distinguish between Facility and Location types and merge them into Location. ERE has a type called Title for titles, honorifics, roles and professions. ACE  has subtypes for entity mentions, which ERE does not. In addition to subtypes, ACE classifies entity mentions into classes (e.g., Specific, Generic and Underspecified), ERE has only Specific. ACE and ERE also have differences in how extents and heads are marked, and levels of entity mentions. 

The purpose of Relation annotation in both ACE and ERE is to extract a representation of the meaning of the text, not necessarily tied    to the underlying syntactic or lexical representation. Both schemes include Physical, Part-Whole, Affiliation and Social relations although the details are a bit different. Both tag relations inside a single sentence and tags only explicit mentions. Nesting of tags is not allowed. Each relation can have up to two ordered Argument slots. Neither model tags negative relations. However, ERE annotates only asserted ("real") events whereas ACE allows others as well, e.g., Believed Events, Hypothetical Events, Desired Events and Requested Events. There is no explicit trigger word in ACE, which annotates the full clause that serves as the trigger for a relation whereas ERE attempts to minimize the annotated span by allowing for the tagging of an optional trigger word or phrase. ACE justifies tagging of each Relation by assigning Syntactic Clauses to them, such as Possessive, PreMod and Coordination. The three types of Relations inn ERE and ACE have sub-types: Physical, Part-Whole, and Social and Affiliation, but ERE collapses ACE types and sub-types to make them more concise, possibly less specific. \cite{Aguilar2014} discuss the similarities and differences between ACE and ERE in detail. 

Events in both ACE and ERE are defined as `specific occurrences' involving `specific participants'. Like entities and relations, ERE is less specific and simplified compared to ACE. Both annotation schemes annotate the same event types: Life, Movement, Transaction, Business, Conflict, Contact, Personnel, and Justice.

\paragraph{RED Annotations}
\cite{Ikuta2014} use another annotation scheme called Richer Event Description (RED), synthesizing co-reference \cite{Pradhan2007,Lee2012} and THYME-TimeML temporal relations \cite{Styler2014}. \cite{Ikuta2014} discusses challenges in annotating documents with the RED schema, in particular cause-effect relations. The usual way to annotate cause-effect relations is using the counter-factual definition of causation in philosophy \cite{Lewis1973,Halpern2005}:
\begin{quote}
``X causes Y" means if X had not occurred, Y would not have happened. 
\end{quote}
However, \cite{Ikuta2014} found that this definition leads to many difficult and sometimes erroneous annotations, and that's why while performing RED annotations, they used another definition \cite{Menzies1999,Menzies2008} which treats causation as ``a local relation depending on intrinsic properties of the events and what goes on between then, and nothing else". In particular, the definition is 
\begin{quote}
``X causes Y" means Y was inevitable given X. 
\end{quote}
In fact, in the annotations performed by \cite{Ikuta2014}, they use the new definition to make judgements, but use the old definition as a precondition to the new one. 

\subsubsection{TAC-KBP Annotations}
The Knowledge Base Population Track (TAC-BKP) was started by NIST in 2009 to evaluate knowledge bases (KBs) 
created from the output of information extraction systems. The primary tasks are a) Entity linking--linking extracted entities to entities in knowledge bases, and b) Slot filling--adding information to entity profiles, information that is missing from the knowledge base \cite{McNamee2010}. Wikipedia articles have been used as reference knowledge bases in evaluating TAC-KBP tasks. For example, given an entity, the goal is to identify individual nuggets of information using a fixed list of inventory relations and attributes. For example, given a celebrity name, the task is to identify attributes such as schools attended, occupations, important jobs held, names of immediate family members, etc., and then insert them into the knowledge base. Many people compare slot filling to answering a fixed set of questions, obtaining the answers and filling in the appropriate slots in the knowledge base. Slot filling in TAC-KBP differs from extraction in ACE and ERE notations in several ways such as TAC-KBP seeks out information for named entities only, chiefly PERs and ORGs, TAC-KBP seeks to obtain values for slots and not mentions, and events are handled as uncorrelated slots, and assessment is like in question-answering. 

Our focus on this paper has been on extracting events, and we know that to extract events properly, we need to explicitly extract event mentions, and also extract associated attributes such as agents, locations, time of occurrence, duration, etc. Rather than explicitly modeling events, TAC-KBP does so implicitly as it captures various relations associated with for example the agent of the event. For example, given a sentence ``Jobs is the founder and CEO of Apple", TAC-KBP may pick "Apple" as the focal entity and identify "Jobs" as the filler of its founder slot, and "Jobs" as the filler of its CEO slot. However, an ACE or ERE annotation program will ideally pick the event as Founding, with Jobs as an argument (say the first argument or arg1, or the Actor) of the event, and "Apple" as another argument, say arg2.

\subsection{Extracting Events}
Many even extraction systems have been built over the years. A big motivator for development of event extraction systems seem to be various contests that are held every few years, although there has been considerable amount of non-contest related research as well. 
Although we discuss extraction of events represented by various formats, the methods are not really different from each other. That is why we discuss TimeML events in more detail and present the others briefly in this section.

\subsubsection{Extracting TimeML Events}
We describe a few of the approaches that have been used for extracting TimeML type events. Quite a few papers that attempt to do so have been published \cite{Sauri2005,Bethard2006,Chambers2007,Llorens2010,Grover2010}, and we pick just a few representative papers.  

\paragraph{The Evita System}: 
\cite{Sauri2005} implemented an event and event feature extraction system called EVITA and showed that a linguistically motivated rule-based system, with some help using statistical disambiguation perfumed well on this task. Evita is claimed to be a unique tool within the TimeML framework in that it is very general, being not based on any pre-established list of event patterns and being domain-independent. Evita can also identify, based on linguistic cues,  grammatical information  associated with event referring expressions, such as tense, aspect, polarity and modality, as stated in the TimeML specification. Evita does not directly identify event participants, but can work with named entity taggers to link arguments to events. 

Evita breaks down the event recognition problem to a number of sub-tasks. Evita preprocesses the input text using the Alembic Workbench POS tagger, lemmatizer to find lexical stems,  and chunkier to obtain phrase chunks, verbal, nominal and adjectival, the three that are commonly used as event referring expressions \cite{Day1997}.  For each subtask after pre-processing, it combines linguistic- and statistically-based knowledge. Linguistic knowledge is used in local and limited contexts such as verb phrases and to extract morphological information. Statistical knowledge is used to disambiguate nominal events. The sub-tasks in event recognition in Evita are: determination of event candidates and then the events, identification of grammatical features of events, additional clustering of event chunks for event detection  and grammatical feature identification in some situations.

For event identification, Evita looks at the lexical items tagged by the preprocessing step. It uses different strategies for identifying events in the three categories: verbs, nouns and adjectives. For identifying events in a verbal chunk, Evita performs lexical look-up and limited contextual parsing in order to exclude weak stative predicates such as {\em be} and generics such as verbs with bare plural subjects. Identifying events expressed by nouns involves a phase of lexical look-up and disambiguation using WordNet, and by mapping events SemCor and TimeBank 1.2 to WordNet synsets. Evita consults 25 subtrees from WordNet where all the synsets denote events. One of these, the largest, is the tree underneath the sunset that contains the word {\em event}. 
If the result of this lexical look-up is not conclusive (i.e., if a nominal occurs as both event and non-event in WordNet), a disambiguation step is applied, based on rules learned by a Bayesian classifier trained on SemCor. 
To identify events from adjectives, Evita uses a conservative approach, where it tags only those adjectives that were annotated as such in TimeBank 1.2, when such adjectives occur as the head of a predicative complement. 

To identify grammatical features (e.g., tense, aspect, modality, polarity and non-finite morphology) of events, Evita uses different procedures based on the part of speech of the event denoting expression. But, in general it involves using morphology, pattern matching, and applying a large number (e.g.,  140 such rules for verbal chunks) simple linguistic rules. However, to identify the event {\em class}, it performs lexical look-up and word sense disambiguation. Clustering is used to identify chunks from the preprocessing stage, that contribute information about the same event, e.g., when some modal auxiliaries and use of copular verbs. Clustering is activated by specific triggers such as the presence of a chunk headed by an auxiliary verb or a copular verb. 

Evaluation of Evita was performed by comparing its performance against TimeBanck 1.2. The reported performance was that Evita had 74.03\% precision, 87.31\% recall and an F-measure of 80.12\% in event detection. Accuracy (precision?) for polarity, aspect and modality was over 97\% in each case. 

\paragraph{Bethard and Martin's approach (2006)}: 
\cite{Bethard2006} use TimeBank-annotated events and identify which words and phrases are events. They consider event identification as a classification task that works on word-chunks. They use the BIO formulation that augments each class label with whether the word is the Beginning, Inside or Outside of a chunk \cite{Ramshaw1995}. 

They use a number of features, categorized into various classes, for machine learning. These include affix features (e.g., three or four characters from the beginning and end of each word), morphological features (e.g., base form of the word, and base form of any verb associated with the word if the word is a noun or gerund, for example), word-class features (e.g., POS tags, which noun or verb cluster a word belongs to where the clusters are obtained using co-occurrence statistics in the manner of \cite{Pradhan2004}), governing features (e.g., governing light verb, determiner type---cardinal or genitive, for example), and temporal features (e.g., a BIO label indicating whether the word is contained inside a TIMEX2 temporal annotation, a governing temporal preposition like since, till, before, etc.). They also use negation features and Wordnet hypernyms as features. For classification, they use the TinySVM implementation of SVM by \cite{Kudo2001}. 

They perform experiments with TimeBank documents using a 90\% stratified sampling for training and 10\% for testing. They obtained 82\% precision and 71\% recall, with an F-measure of 0.759. They did compare their algorithm with an version of Evita they programmed themselves; this system obtained 0.727 F-measure, and thus Bethard and Martin's approached performed about 4\% better. When Bethard and Martin's system was extended to identifying semantic class of an event, it did not perform as well, obtaining precision of 67\%, recall of 51\%, and F-measure of 0.317. However, the system was much better at identifying the classes of verbs with F-measure of 0.707 compared to finding classes of nouns with an F-measure of 0.337 only. 

\paragraph{Llorens' et al.'s approach:}
TIPSem (Temporal Information Processing based on Semantic information) is  a system  that participated in the TemEval-2 Competition \cite{Verhagen2010} in 2010, which  presented several tasks to participants, although we are primarily  interested in the event extraction task.   
TIPSem achieved the best F1 score in all the tasks in TempEval-2  for Spanish, and for English it obtained the best F1 metric in the task of extracting events, which required  the recognition and classification of events as defined by TimeML EVENT tag. 

TIPSem learns Conditional Random Field (CRF) models using features for different language analysis levels, although the approach focuses on semantic information, primarily semantic roles and semantic networks.
Conditional Random Fields present a popular and efficient machine learning technique for supervised sequence labeling \cite{Lafferty2001}.  

The features used for training the CRF models are similar to one used by others such as Bethard and Martin, although details vary. However, they add semantic role labels to the mix of features. In particular, they  identify roles for each governing verb. 
Semantic role labeling \cite{Gildea2002,Moreda2007,Punyakanok2004} identifies for each predicate in a sentence, semantic roles and determine their arguments (agent, patient, etc.) and their adjuncts (locative, temporal, etc.). The previous two features were combined in TIPSem to capture the relation between them. The authors think this combination introduces additional information by distinguishing roles that are dependent on different verbs. The importance of this falls especially on the numbered roles (A0, A1, etc.) meaning different things when depending on different verbs.

The test corpus consists of 17K words for English and 10K words for Spanish, provided by the organizers of TempEval-2. For English, they obtained precision of 0.81, recall of 0.86 and F-measure of 0.83 for recognition  with event classification accuracy of 0.79; for Spanish the numbers were 0.90, 0.86, 0.88 for recognition and 
0.66 for classification accuracy. We provide these numbers although we know that it is difficult to compare one system with another, for example Bethard and Martin's system with TIPSem since the corpora used are difference. 

\paragraph{TempEval-3: 2012}
As in TempEval-2, TempEval-3 \cite{UzZaman2012} participants took part in a task where they had to determine the extent of the events in a text as defined by the TimeML EVENT tag. In addition, systems may determine the value of the features CLASS, TENSE, ASPECT, POLARITY, MODALITY and also identify if the event is a main event or not. The main attribute to annotate is CLASS.

The TempEval-3 dataset was mostly  automatically generated, using a temporal merging system. The half-million token text corpus from English Gigaword2 was automatically annotated  using TIPSem, TIPSem-B \cite{Llorens2010} and TRIOS \cite{UzZaman2010}. These systems were re-trained on the TimeBank and AQUAINT corpus, using the TimeML temporal relation set. The outputs of these three state-of-the-art system were merged using a merging algorithm \cite{UzZaman2012}.  The dataset used comprised about 500K tokens of ``silver" standard data and about 100K tokens of ``gold" standard data for training, compared to the corpus of roughly 50K tokens corpus used in TempEval 1 and 2.

There were seven participants and all the participants except one used machine learning approaches. The top performing system was ATT-1  \cite{Jung2013} with precision 81.44, recall 80;67 and F1 of 81.05 for event recognition, and 71.88 for event classification. Close behind was the ATT-2 system \cite{Jung2013} with precision, recall and F-1 of 81.02, 80.81 and 80.92 for event recognition respectively, and 71.10 for event classification. Both systems used MaxEnt classifiers with 

Obviously, different sets of features  impact on the performance of event  recognition and classification \cite{Adafre2005,Angeli2012,Rigo2011}. In particular, \cite{Rigo011} also examined performance based on different sizes of n-grams in a small scale (n=1,3).
Inspired by such work, 
in building the ATT systems, the creators intended to systematically investigate the performance of various models and  for each  task, they trained twelve models exploring these two dimensions, three of which we submitted for TempEval-3, and of these three performed among the top ten in TempEval-3 Competition. 

The  ATT-1 models include lexical, syntactic and semantic features, ATT-2 models include only lexical and syntactic features, and ATT-3 models include only lexical features, i.e., words. They experimented with context windows of 0, 1, 3, and 7 words preceding and following the token to be labeled. For each window size, they trained ATT-1, ATT-2 and ATT-3 models. The ATT-1 models had 18 basic features per token in the context window for up to 15 tokens, so up to 270 basic feaures for each token to be labeled. The ATT-2 models had 16 basic features per token in the context window, so up to 240 basic features for each token to be labeled. The ATT-3 models had  just 1 basic feature per token in the context window, so up to 15 basic features for each token to be labeled.

For event extraction and classification,  and event feature classification, they used the  efficient binary MaxEnt classifier for multi-class classification, available in  the machine learning toolkit LLAMA \cite{Haffner2006}. They also used  LLAMA's pre-processor to build unigram, bigram and trigram extended features from basic features.  

For event and time expression extraction, they trained BIO classifiers.  
It was  found that  the absence of semantic features causes only small changes in F1. The absence of syntactic features causes F1 to drop slightly (less than 2.5\% for all but the smallest window size), with recall decreasing while precision improves somewhat. F1 is also impacted minimally by the absence of semantic features, and about 2-5\% by the absence of syntactic features for all but the smallest window size.1

A was  surprising that  that  ATT-3  models that use words only performed well, especially in terms of precision (precision, recall and F2 of  81.95,  75.57 and 78.63  for event recognition, and 69.55 F1 for event classification) . It is also surprising that the words only models with window sizes of 3 and 7 performed as well as the models with a window size of 15. These results are promising for ``big dataÓ text analytics, where there may not be time to do heavy preprocessing of input text or to train large models.

\subsubsection{Extracting Events Using Other Representations}
We have already discussed several approaches to extraction of events represented by  TimeML representation. Extracting events that use other representation is not very different, but different representations have existed and exist, and therefore we briefly present some such attempts. Some of these predate the time TimeML became popular. 
For example, the various Message Understanding Conferences (MUCs, seven were organize by DARPA from 1987 to 1997), asked participants to extract a small number of relations and events. For instance, MUC-7,  the last one, called for the extraction of 3 relations {\em (person-employer, maker-product, and organization-location)} and 1 event {\em spacecraft launches}. 

For example, the MUC-7 and ACE events did not attempt to cover all events, but a limited number of pre-specified event types or classes that participants need to detect during a contest period, based on which the contestants submit papers for publication. The number and the type of arguments covered are also limited and are pre-specified before the competitions start.  

\paragraph{The REES System:}

\cite{Aone2000} discuss a  relation and event extraction system covering  areas such as political, financial, business, military, and life-related topics.  The system consists of  tagging modules, a  co-reference resolution module, and a temple generation module. They store the events generated in MUC-7\cite{Chinchor1998}  format, which is not very unlike the ACE format. 

Events are extracted along with their event participants, e.g., who did what to whom when and where? For example, for a BUYING event, REES extracts the buyer, the artifact, the seller, and the time and location of the BUYING event. REES  covers 61 types of events. There are 39 types of relations. 

The tagging component consists of three modules: NameTagger, NPTagger and EventTagger. Each module relies on the same pattern-based extraction engine, but uses different sets of patterns. The NameTagger recognizes names of people, organizations, places, and artifacts (only vehicles in the implemented system).
The NPTagger then takes the output of the NameTagger and first recognizes non-recursive Base Noun Phrase (BNP) \cite{Ramshaw1995}, and then complex NPs for only the four main semantic types of NPs, i.e., Person, Organization, Location, and Artifact (vehicle, drug and weapon). The EventTagger recognizes events applying its lexicon-driven, syntactically-based generic patterns. 

REES uses a declarative, lexicon-driven approach. This approach requires a lexicon entry for each event-denoting word, which is generally a verb. The lexicon entry specifies the syntactic and semantic restrictions on the verb's arguments.  After the tagging phase, REES sends the  output through a rule-based co-reference resolution module that resolves: definite noun phrases of Organization, Person, and Location types, and singular personal pronouns. REES outputs the extracted information in the form of either MUC-style templates or XML.

One of the challenges of event extraction is to be able to recognize and merge those event descriptions which refer to the same event. The Template Generation module uses a set of declarative, customizable rules to merge co- referring events into a single event.

The  system's recall, precision, and F-Measure scores for the training set (200 texts) and the blind set (208 texts) from about a dozen news sources. On the so-called training set, the system achieved F-measure of 64.75 for event extraction and 75.35 for relation extraction. 
The blind set F-Measure for 31 types of relations (73.95

\paragraph{Ahn's approach}
As seen earlier in Subsection \ref{subsection:ACEEvents}, the way ACE events are specified, they have   a lot of details that need to be extracted.  \cite{Ahn2006} follows several steps to extract events and uses machine learning algorithms at every step.  The steps are pre-processing of text data, identifying anchors, assigning  event types, extracting 
 arguments
 identifying attributes of events such as modality, polarity, genericity and tense, and finally
identifyings event co-referents of the same individuated event.
In other words, Ahn attempts to cover all the steps sequentially, making the simplifying assumption that they are unrelated to each other. 

A single place  in a textual document which may be considered the primary place of reference or discussion about an event is called the event anchor. Ahn treats finding the anchor for an event within a document as a {\em word classification task}, using a two-stage classification process. He uses a binary classifier to classify a word as being an event anchor or not. He then classifies those identified as event anchors into one of the event classes. Ahn used one classifier for binary classification and then another classifiers to classify only the positive instances. 

Ahn treats identifying event arguments as a pair classification task. Each event mention is paired with each of the entity, time and value mentions occurring in the same sentence to form a single classification instance. There were 35 role types in the ACE 2006 task, but no event type allows arguments of all types. Each event type had its own set of allowable roles. The classification experiment run was a multi-class classification where a separate multi-class classifier was used for each event type. 
Ahn trains a separate classifier for each attribute. Genericity, modality, and polarity are each binary classification tasks, while tense is a multi-class task.
For event coreference, Ahn follows the approach given in \cite{Florian2004}.  Each event mention in a document is paired with every other event mention, and a classifier assigns to each pair of mentions the probability that the paired mentions corefer. These probabilities are used in a left-to-right entity linking algorithm in which each mention is compared with all already-established events (i.e., event mention clusters) to determine whether it should be added to an existing event or start a new one. 

Ahn experimented with various combinations of a maximum entropy classifier MegaM \cite{Daume2004}  and   a memory-based nearest neighbor classifier  called TIMBL \cite{Daelemans2004}, for the various tasks.

The ACE specification provided a way to measure the performance of an event extraction system. The evaluation called ACE value is obtained by scoring each of the component tasks individually and then obtaining a normalized summary value. 
Overall, using the best learned classifiers for the various subtasks, they achieve an ACE value score of 22.3\%, where the maximum score is 100\%. The value is low, but other systems at the time had comparable performance. 

\paragraph{Naughton 2008:}
\cite{Naughton2008} describe an approach to classify sentences in a document as specifying one or more events from a certain ACE 2006 class. They classify each sentence in a document as containing an instance of a certain type or not. 
  Unlike \cite{Ahn2006}, they are not interested in identifying arguments or any additional processing. Also, unlike Ahn who classifies {\em each word} as possibly being an event anchor for a specific type of ACE event, Naughton et al.  perform a classification of {\em each sentence} in a document as being an {\em on-event sentence} or an {\em off-event sentence}. An {\em on-event} sentence is a sentence that contains one or more instances of the target event type. An {\em off-event} sentence is a sentence that does not contain any instances of the target event type.  They use several approaches to classify a sentence as on-event or off-event. These include the following: SVM-based machine learning  \cite{Joachims1998}, language modeling approaches using count smoothing, and 
a manual  approach which looks for Wordnet synonyms or hypernyms of certain trigger words in a sentence. 
 
 Naughton et al.  found that 1) use of a large number of features to start but then reduction of these features using information gain, and 2) use of SVM produces the best results although all versions of SVM (i.e., with all features with no reduction, just the terms without complex features, or a selection of terms and other features) all work very well. A ``surprising" finding was that the ``manual" trigger-based classification approach worked almost as well as the SVM based approaches. 

\subsection{Determining Event Coreference}
When an event is mentioned in several places within a document, finding which references are to the same event is called determining event coreference. These are co-referents to the event. Determining when two event mentions in text talk about the same event or co-refer is a difficult problem. As \cite{Hovy2013} point out that the events may be actual occurrences or hypothetical events. 

\subsubsection{Florian's Approach to Coreference Resolution}
\cite{Florian2004}  present a statistical language-independent framework for identifying and tracking named, nominal and pronominal references to entities within unrestricted text documents, and chaining them into groups corresponding to each logical entity present in the text. The  model  can use arbitrary feature types, integrating a variety of lexical, syntactic and semantic features. The mention detection model  also uses feature streams derived from different named entity classifiers. 

For mention detection, the approach used is  based on a log-linear Maximum Entropy classifier \cite{Berger1996} and  a linear Robust Risk Minimization classifier  \cite{Zhang2002}. Then they use a MaxEnt model for predicting whether a mention should or should not be linked to an existing entity, and  to build entity chains. 
Both classifiers can integrate arbitrary types of information and are converted into suitable for sequence classification for both tasks.

For entity tracking, the process works from left to right. It starts with an initial entity consisting of the first mention of a document, and the next mention is processed by either linking it with one of the
existing entities, or starting a new entity. 
Atomic features used by the entity linking algorithm include 
string match, context, 
mention count,  distance between the two mentions in words and sentences,  editing distance, properties of pronouns such gender, number and reflexiveness.
The best combination of features was able to obtain slightly more than 73\% F-1 value using both RRM and MaxEnt algorithms for mention detection. 

Entity tracking was evaluated in terms of what is called the ACE value. A gauge of the performance of an EDT system is the ACE value, a measure developed especially for this purpose. It estimates the normalized weighted cost of detection of specific-only entities in terms of misses, false alarms and substitution errors. Florian et al. achieved an ACE value of 73.4 out of 100 for the MaxEnt classifier and 69.7 for the RRM classifier. 

\cite{Ahn2006} follows the approach by \cite{Florian2004} for entity coreference determination. He uses a binary classifier to determine if any two event mentions in the document refer to the same event. Thus, he pairs each event with every other event, and the classifier assigns each pair a probability that they are the same. The probability is used with entity linking/matching algorithm to determine event co-reference. Event co-referencing requires event mentions to be clustered to event clusters. Event mentions in a cluster are the same event. The system described here obtained an ACE value of between 88-91\%, where the maximum ACE value is 100\%. 

\cite{Ahn2006} uses the following features for event co-reference determination. Let the candidate be the earlier event mention and the anaphor be the later mention. 
\begin{itemize}
\item The anchors for the candidate and the anaphor, the full or original form, and also in lowercase, and POS tag. 
\item Type of the candidate event and the anaphor event.
\item Depth of candidate anchor word in parse tree. 
\item Distance between the candidate and anchor, measured in sentences. 
\item Number, heads, and roles of shared arguments, etc.
\end{itemize}

\subsubsection{Bejan and Harabagiu}

Supervised approaches to solving event coreference use  linguistic properties  to decide if a pair of event mentions is coreferential \cite{Humphreys1997,Bagga1999,Ahn2006,Chen2009}. These  models depend on labeled training data, and annotating a large corpus with event coreference information requires substantial manual effort. In addition, since these models make local pairwise decisions, they are unable to capture a global event distribution at topic or document collection level. \cite{Bejan2010}
present  how nonparametric Bayesian models can be  applied to an open-domain event coreference task in an unsupervised manner.  

The first model extends the hierarchical Dirichlet process \cite{Teh2006} to take into account additional properties associated with  event mentions. The second model overcomes some of the limitations of the first model, and uses the infinite factorial hidden Markov model \cite{Gael2009} coupled to the infinite hidden Markov model \cite{Beal2001} in order to consider a potentially infinite number of features associated with observable objects which are event mentions here, perform an automatic selection of the most salient features, and  capture the structural dependencies of observable objects  or event mentions at the discourse level. Furthermore, both models can work with a potentially infinite number of categorical outcomes or events in this case. 

Two event mentions  corefer if they have the same event properties and share the same event participants.
To find coreferring event mentions, Bejan and Harabagiu describe  words that may be possible event mentions with  lexical features, class features such as POS and event classes such \cite{Pustejovsky2003} as occurrence, state and action, Wordnet features, semantic features obtained by a semantic parse \cite{Bejan2007} and the predicate argument structures encoded in PropBank annotations \cite{Palmer2005} as well as semantic annotations encoded in the FrameNet corpus \cite{Baker1998}.

The first model represents each event mention by a finite number of feature types, and is also inspired by the Bayesian model proposed by \cite{Haghighi2007}. 
In this model, a Dirichlet process (DP) \cite{Ferguson1973} is associated with each document, and each mixture component (i.e., event) is shared across documents since 
In the process of generating an event mention, an event index z is first sampled by using a mech- anism that facilitates sampling from a prior for in- finite mixture models called the Chinese restaurant franchise (CRF) representation, as reported in \cite{Teh2006}. 

The second model they use is called the iHMM-iFHMM model (infinite hidden Markov model--infinite factorial hidden Markov model). The iFHMM framework uses the Markov Indian buffet process (mIBP) \cite{Gael2009} in order to represent each object as a sparse subset of a potentially unbounded set of latent features \cite{Ghahramani2005,Gael2008}, Specifically, the mIBP defines a distribution over an unbounded set of binary Markov chains, where each
chain can be associated with a binary latent feature that evolves over time according to Markov
dynamics.
The iFHMM allows a  flexible representation of the latent structure by letting the number of parallel Markov chains  be learned from data, it cannot be used  where the number of clustering components  is infinite. An iHMM represents a nonparametric extension of the hidden Markov model (HMM) \cite{Rabiner1989} that allows performing inference on an infinite number of states. To further increase the representational power for modeling discrete time series data, they develop a nonparametric extension that combines the best of the two models, and lets the  two parameters M and K be learned from data
Each step in the new iHMM-iFHMM generative process is performed in two phases: (i) the latent feature variables from the iFHMM framework are sampled using the mIBP mechanism; and (ii) the features sampled so far, which become observable during this second phase, are used in an adapted version of the beam sampling algorithm 
\cite{Gael2008a} to infer the clustering components (i.e., latent events).

They report  results in terms of recall (R), precision (P), and F-score (F) by employing the mention-based B3 metric \cite{Bagga1998}, the entity-based CEAF metric \cite{Luo2005}, and the pairwise F1 (PW) metric. 
Their experiments for show that both of these models work well when the feature and cluster numbers are treated as free parameters, and the selection of feature values is performed automatically.

\subsubsection{Hovy et al} 
\cite{Hovy2013} argue that events represent complex phenomena  and can therefore co-refer {\em fully}, being identical,  like other researchers have discussed, or co-refer {\em partially}, being quasi-identical or only partially identical. Two event mentions fully co-refer if their activity, event or state representation is identical in terms of all features used (e.g., agent, location or time). Two event mentions are quasi-identical if they partially co-refer, i.e., most features are the same, but there may be additional details to one or the other. 

When two events fully co-refer,  Hovy et al.  state they may be lexically identical (i.e., the same senses of the same word, e.g., {\em destroy} and {\em destruction}), synonymous words, one mention is a wider reading of the other (e.g., {\em The attack took place yesterday} and {\em The bombing killed four people}), one mention is a paraphrase of the other with possibly some syntactic differences (e.g., {\em He went to Boston} and {\em He came to Boston}), and one mention deictically refers to the other (e.g., {\em the party} and {\em that event}).  Quasi-identity or partial co-reference may arise in two ways:  {\em membership identity} or {\em subevent identity}. Membership identity occurs when one mention, say A, is a set of multiple instances of the same type of event, and the other mention, say B, is one of the individual events in A (e.g., {\em I attended three parties last week. The first one was the best.}). Subevent identity is found when one mention, say A, is a stereotypical sequence (or script) of events whereas the other mention, say B, is one of the actions or events within the script (e.g., {\em The family ate at the restaurant. The dad paid the waitress at the end.}). 

Hovy et al.attempt to build a corpus containing event co-reference  links with high quality annotations, i.e., annotations with high inter-annotator agreement, to be useful for machine learning. They have created two corpora to assist with a project on automated deep reading of texts. One corpus is in the domain of violent events (e..g., bombings, killens and wars), and the other one containing texts about the lives of famous people. In both of these corpora, they have annotated a limited number of articles with full and partial co-references. 

\subsubsection{Delmonte} 
\cite{Delmonte2013} claims that performing event co-reference with high accuracy requires deep understanding of the text and statistically-based methods, both supervised and unsupervised, do not perform well. He claims that this is the case because because it is absolutely necessary to identify arguments of an event reliably before event co-references can be found. Arguments are difficult to identify because many are implicit and linguistically unexpressed. 
Successful even co-reference identification needs determination of spatio-temporal anchoring and locations in time and space are also very often implicit. 

The system he builds uses a linguistically based semantic module, which has  a number of different submodules which take care of Spatio-Temporal Reasoning, Discourse Level Anaphora Resolution, and determining Topic Hierarchy.
 The coreference algorithm works as follows: for each possible referent it check all possible coreference links, at first using only the semantic features, which are: wordform and lemma identity; then semantic similarity measured on the basis of a number of similarity criteria which are lexically based. The system searches WordNet synsets and assign a score according to whether the possible referents are directly contained in the same synset or not. A different score is assigned if their relation can be inferred from the hierarchy. Other computational lexical resources they use  include FrameNet and Frames hierarchy; SumoMilo and its semantic classification.
 
 After collecting all possible coreferential relations, the system  filters out those links that are inconsistent or incompatible. 
Argument structure and spatiotemporal relations are computed along with dependence relations; temporal logical relations as computed using an adaptation of Allen's algorithm. 
The system also computes semantic similarity, where high values are preferred.  
 The paper does not give any results to support the initial hypothesis, although the ideas are interesting. 
 
\subsubsection{Cybulska and Vossen}

\cite{Cybulska2015}  use granularity in computing  event coreference. 
The intuition  is, that an event with a longer duration, that
happens on a bigger area and with multiple particpants  (for instance {\em a war between Russia and Ukraine}) might be related to but will probably not fully corefer with a Òlower levelÓ event of shorter duration and with single participants involved (e.g. {\em A Russian soldier has shot dead a Ukrainian naval officer}).

Coreference between mentions of two events is determined by computing compatibility of contents of event attributes. The attributes used  are  event trigger, time,  location, human and non-human participant slots \cite{Cybulska2014a}. 
Granularity size is mentioned in terms of durations of event actions \cite{Gusev2011} and granularity levels of event participants, time and locations. 
Granularity is given in terms of partonomic relations or through the part-of relation, between entities and events, using the taxonomy of meronymic relations by \cite{Winston1987}. Granularity levels of the human participant slot are contained within WinstonÕs et al. Member-Collection relations. The temporal granularity levels make part of WinstonÕs Portion-Mass relationships and locational levels are in line with Place-Area relations in WinstonÕs taxonomy.

Cybulska and Vossen experimented with a decision-tree supervised pairwise binary classifier to determine coreference of pairs of event mentions. They also ran experiments with a linear SVM and a multinomial Naive Bayes classifier but the decision-tree classifier outperformed both of them. 

For the experiments, Cybulska and Vossen use the ECB+ dataset \cite{Cybulska2014b}. 
The ECB+ corpus contains a new corpus component, consisting of 502 texts, describing different instances of event types.
They provide results in terms of several metrics: recall, precision  and F-score,  MUC \cite{Vilain1995}, B3 \cite{Bagga1998}, mention-based CEAF \cite{Luo2005}, BLANC \cite{Recasens2011}, and CoNLL F1 \cite{Pradhan2011}, and find that the introduction of the granularity concept into similarity computation improves results for every metric.

\section{Biomedical Event Extraction}
Researchers are interested in extracting information from the huge amount of biomedical literature  published on a regular basis. Of course, one aspect of information extraction is event extraction, the focus of this paper. In the biomedical context, an event extraction system tries to extract details of bimolecular interactions among biomedical entities such as proteins and genes, and the processes they take part in, as described in terms of textual documents. Manually annotated corpora are used to train machine learning  techniques and  evaluate event extraction techniques. 

There have been several workshops on biomedical natural language processing. We focus on the BioNLP Shared Tasks in recent years that had competitions on event extraction. 
There have been three BioNLP Shared Task competitions so far: 2009, 2011, and 2013. 
 The BioNLP 2009 Shared Task  \cite{Kim2009} was based on the GENIA corpus \cite{Kim2003} which contains PubMed\footnote{\url{http://www.ncbi.nlm.nih.gov/pubmed}} abstracts of articles on transcription factors in human blood cells. 
There was a second BioNLP Shared Task competition organized in 2011 to measure the advances in approaches and associated results \cite{Kim2011}. The third BioNLP ST was held in 2013.  We discuss some notable systems from BioNLP ST 2011 and 2013.

Before the  BioNLP Shared Tasks, event extraction in the biomedical domain usually classified each pair of named entities (usually protein names) co-occurring in the text as interacting or not. BioNLP Shared Tasks extended such an approach by adding relations such as {\em direction}, {\em type} and {\em nesting}. An event  defines the type of interaction, such as {\em phosphorylation}, and is usually marked in the text with a {\em trigger word} (e.g., {\em phosphorylates}) describing the interaction. This word forms the core of the event description. A directed event has roles that have inherent directionality such as {\em cause} or {\em theme}, the agent or target of the biological process. In addition, events can act as arguments of other events, creating complex nested structures. For example, in the sentence {\em Stat3 phosphorylation is regulated by Vav}, a {\em phosphorylation}-event is the argument of the {\em regulation}-event. 

The BioNLP Shared Tasks  provide task definitions, benchmark data and evaluations, and participants compete by developing systems to perform the specified tasks. 
The theme of BioNLP-ST 2011 was a generalization of the 2009 contest, generalized in three ways: text types, event types, and subject domains. The  2011 event-related tasks were arranged in four tracks:
 GENIA task (GE) \cite{Kim2011},  Epigenetics and Post-translational Modifications (EPI) \cite{Ohta2011},  Infectious Diseases  (ID)\cite{Pyysalo2011}, and the Bacteria Track \cite{Bossy2011,Jourde2011}.

Of the  four event-related shared tasks in BioNLP 2011, the first three were related to event extraction. 
The Genia task was focused on the domain of transcription factors in human blood cell. Trascription is a complex but just the first step in the process in which the instructions contained in the DNA in the nucleus of a cell are used to produce proteins that control most life processes. Transcription factors are proteins that control the transcription process. The EPI task was focused on events related to epigenetics, dealing with protein and DNA modifications, with 14 new event types, including major protein modification types and their reverse reactions. Epigenesis refers to the development of a plant or animal from a seed, spore or egg, through a sequence of steps in which cells differentiate and organs form.  The EPI task was designed toward pathway extraction and curation of domain databases \cite{Wu2003,Ongenaert2008}. A biological pathway refers to a sequence of actions among molecules in a cell that leads to a certain product or a change in the cell. The ID task was focused on extraction of events relevant to biomolecular mechanisms of infectious diseases from full length publications. 
Tasks other than ID focused on abstracts only. 

In this paper, we discuss the systems and approaches for  only the 2011 GE Task. This is because several of the winning systems for the GE Task did well in the other two relevant tasks as well. The Genia Task  is described in Table~\ref{table:BioNLP2011GeniaTask}.  The table shows for each event type, the primary and secondary arguments to be extracted. For example, a {\em phosphorylation} event is primarily extracted with the protein to be phosphorylated, which is the addition of a phosphate group to a protein or other organic molecule. As secondary information, the specific site to be phosphorylated may be extracted. From a computational viewpoint, the event types represent different levels of complexity. When only primary arguments are considered, the first five event types in Table \ref{table:BioNLP2011GeniaTask} are classified as {\em simple events}, requiring only unary arguments. The {\em binding} and {\em regulation} types are more complex. Binding requires the detection of an arbitrary number of arguments, and Regulation requires detection of recursive event structure. 

\begin{table}
\begin{tabular}{| l | p{1.85in}  | l | l |} \hline
{\bf Event Type} & {\bf Primary Argument } & {\bf Secondary Argument }\\ \hline \hline
Gene expression & Theme (Protein) & \\ 
Transcription & Theme (Protein) & \\ 
Protein catabolism & Theme (Protein) & \\ 
Phosphorylation & Theme (Protein) & Site(Entity)\\ 
Localization & Theme (Protein) & AllLoc(entity), ToLoc (Entity) \\ \hline
Binding & Theme (Protein)+ & Site(Entity)+\\ \hline
Regulation & Theme (Protein/Event), Cause (Protein/Event) & Site(Entity), CSite(Entity) \\ \hline
Positive regulation & Theme (Protein/Event), Cause (Protein/Event) & Site(Entity), CSite(Entity) \\ \hline
Negative regulation & Theme (Protein/Event), Cause (Protein/Event) & Site(Entity), CSite(Entity)\\ \hline
\end{tabular}
\caption{Event types and their arguments in the Genia event task. The type of arguments, primary and secondary, to be extracted from the text, are also given.}
\label{table:BioNLP2011GeniaTask}
\end{table}

Consider the sentence {\em In this study we hypothesized that the phosphorylation of TRAF2 inhibits binding to the CD40 cytoplasmic domain.} Here there are two protein  (entity) names:  {\em TRAF2} and {\em CD40}. The word {\em phosphorylation} refers to an event; this string is a trigger word. Thus, the goal of 
the GE task was to identify  a structure like the ones in Tables \ref{table:BioNLP2011EventStructure1}  and \ref{table:BioNLP2011EventStructure2} . In the tables, $T_i$ represents a trigger word, and $E_i$ represents an event associated with the corresponding trigger word. There are three events, $E_1$ is the {\em phosphorylation} event, $E_2$ is the {\em binding} event and $E_3$ is the {\em negative regulation} event. For each trigger word, we see the starting and ending character positions in the entire string. For each event, we see the participants in it. The second task identifies an additional {\em site} argument. 

\begin{table}
\begin{center}
	\includegraphics[scale=0.6]{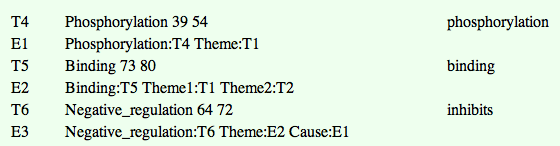}
\end{center}
\caption{Structure of an Event in BioNLP 2011 Contest,  corresponding to Task 1 (GE) }
\label{table:BioNLP2011EventStructure1}
\end{table}

\begin{table}
\begin{center}
	\includegraphics[scale=0.6]{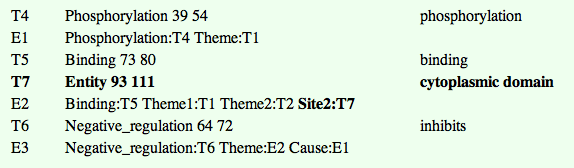}
\end{center}
\caption{Structure of an Event in BioNLP 2011 Contest,  corresponding to Task 2 (EPI)}
\label{table:BioNLP2011EventStructure2}
\end{table}

Table \ref{table:BioNLP2011BestResults} shows the best results for various tasks in the BioNLP 2011 contests. \cite{Kim2011} note an improvement of 10\% over the basic GE task, in 2011 (Task GEa), compared to 2009. The results of the GE tasks show that automatic extraction of simple events--those with unary arguments, e.g., gene expression, localization and phosphorylation---can be achieved at about 70\% in F-score, but the extraction of complex events, e.g., binding and regulation is very challenging, with only 40\% performance level. 
The GE and ID results show that generalization to full papers is possible, with just a small loss in performance. The results of {\em phosphorylation} events in GE and EP are similar (GEp vs. EPIp), which leads \cite{Kim2011} to conclude that removal of the GE domain specificity does not reduce event extraction performance by much. EPIc results indicate that there are challenges to extracting similar event types that need to be overcome; EPIf results indicate that there are difficult challenges in extracting additional arguments. The complexity of the ID task is similar to that of the GE task; this shows up in the final results, also indicating that it is possible to generalize to new subject domains and new argument (entity) types. 

\begin{table}
\begin{center}
\begin{tabular}{| l | l |}\hline
Task & Evaluation Results \\ \hline
BioNLP\_ST 2000 & 46.73 / 58.48/ 51.95\\ 
Miwa et al. (2010b) & 48.62 / 58.96 / 53.29\\
LLL 2005 (LLL) & 53.00 / 55.60 / 54.30 \\ \hline
GE abstracts (GEa) & 50.00 / 67.53 / 57.46 \\ 
GE full texts (GEf) & 47.84 / 59.76 / 53.14 \\
GE PHOSPHORYLATION (GEp) & 79.26 / 86.99 / 82.95 \\
GE LOCALIZATION (GEl) & 37.88 / 77.42 / 50.87 \\ \hline
EPI full task (EPIf) & 52.69 / 53.98 / 53..3 \\
EPI core task (EPIc) & 68.51 / 69.20 / 68.86 \\
EPI PHOSPHORYLATION (EPIp) & 86.15 / 74.67 / 80.00 \\ \hline
ID full task (IDf) & 48.03 / 65.97 / 55.59 \\
ID core task (IDc) & 50.62 / 66.06 / 57.32 \\ \hline
\end{tabular}
\end{center}
\caption{Best results for various sub-tasks in BioNLP\_ST 2011. Recall/precision/F-score \%}
\label{table:BioNLP2011BestResults}
\end{table}

Below, we provide a brief description of some of the approaches to biomedical event extraction from the BioNLP 2011 contests.

\subsection{Technical Methods Used in BioNLP Shared Tasks 2011}
The team that won the GE Task  was the FAUST system \cite{Riedel2011a}, followed by the UMass system \cite{Riedel2011b}, then the UTurku system \cite{Bjorne2011}. The performance of these three systems on the various tasks is given in Table \ref{table:BioNLP2011GeniaTaskEvaluation}. In addition, we have the Stanford system in the table because it performed fairly well on the tasks.

\begin{table}
\begin{tabular}{| l l | l l l l |}\hline
\multicolumn{2}{|c|}{Team} & Simple Event & Binding & Regulation & All\\ \hline \hline
\multirow{3}{*} {FAUST}  & W & 68.5/80.3/73.9&44.2/53.7/48.5  & 38.0/54.9/44.9 &  49.4/64.8/56.0\\
& A & 66.2/81.0/72.9& 45.5/58.1/51.1 & 39.4/58.2/47.0 & 50.0/67.5/57.5 \\
& F & 75.6/78.2/76.9 & 41.1/44.7/ 42.8 & 35.0/48.2/40.6 & 47.9/58.5/52.7 \\ \hline 
\multirow{3}{*}{UMass} & W & 67.0/ 81.4/ 73.5 & 43.0/ 56.4/ 48.8 & 37.5/52.7/43.8 & 48.5/64.1/55.2\\
& A & 64.2/80.7/71.5 & 43.5/60.9/50.8 & 38.8/55.1/45.5 & 48.7/65.9/56.1 \\
& F & 75.6/83.1/79.2 & 41.7/47.6/44.4 & 34.7/47.5/40.1 & 47.8/59.8/53.1\\ \hline
\multirow{3}{*}{UTurku} & W &  68.2/76.5/72.1 & 42.8/43.6/43.3 & 38.7/47.6/42.7 & 49.6/57.7/53.3 \\
& A & 65.0/76.7/70.4 & 45.2/50.0/47.5 & 40.4/49.0/44.3 & 50.1/59.5/54.4\\
& F &  78.2/75.8/77.0 & 37.5/31.8/34.4 & 35.0/44.5/39.2 & 48.3/53.4/50.7\\ \hline
\multirow{3}{*}{Stanford} & W & 65.8/76.8/70.9 & 39.9/9.9/44/3 & 27.6/48.8/35.2 & 42.4/61.1/50.0\\
& A & 62.1/77.6/69.3 & 42.4/54.2/47.6 & 28.3/50.0/36.1 & 42.6/62.7/50.7\\
& F & 75.6/75.0/75.3 & 34.0/40.2/36.9 & 26.0/46.1/33.3 & 41.9/57.4/48.4\\ \hline
\end{tabular}
\caption{Evaluation results (recall / precision / f-score for Task 1 in Whole data set (W), Abstracts only (A) and Full papers only (F) }
\label{table:BioNLP2011GeniaTaskEvaluation}
\end{table}

\begin{figure}
	\centering
	\includegraphics[scale=0.9]{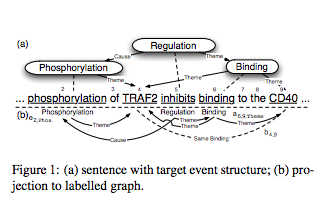}
	\caption{(a) Sentence with target event structure, (b) Projection to labeled graph. ***Redraw the image***}
	\label{fig:Riedel2011EventStructure}
\end{figure}

The UMass system \cite{Riedel2011b} looks at a sentence as having an event structure, and then projects it onto a labeled graph. See Figure \ref{fig:Riedel2011EventStructure} for a target event structure and the projected graph for the sentence fragment {\em Phosphorylation of TRAF2 inhibits binding to CD40}. The system searches for a structure that connects the event and its participating entities and imposes certain constraints on the structure. Thus, the UMass system treats the search for such a structure as an optimization problem. To formulate this optimization problem, the system represents the structure in terms of a set of binary variables, inspired by the work of \cite{Riedel2009,Bjorne2009}. These binary variables are based on the projection of the events to the labeled graph. An example of a binary variable is $a_{i,l.r}$ to indicate that between positions $i$ and $l$ in the sentence, there is an edge labeled $r$ from a set of possible edge labels $R$. Another such binary variable is $t_{i,p,q}$ that indicates that at position $i$, there is a binding event with arguments $p$ and $q$. Given a number of such variables, it is possible to write an objective function to optimize in order to obtain events and entity bindings. 
The system
decomposes the biomedical event extraction task into three sub-tasks: (a) event triggers and outgoing edges on arguments, (b) event triggers and incoming edges on arguments, and (c) and protein-protein bindings. The system obtains an objective function for each of the sub-tasks. It solves the three optimization problems one by one in a loop, till no changes take place, or up to a certain number of iterations. 
The approach uses  optimizing by dual decomposition \cite{Komodakis2007,Rush2010} since the dual of the original optimization problem is solved. 

The Stanford system \cite{McClosky2011} exploits the observation that event structures bear a close relation to dependency graphs \cite[Chapter 12]{Jurafsky2009}. They cast bimolecular events in terms of these structures which are pseudo-syntactic in nature. They claim that standard parsing tools such as maximum-spanning tree parsers and parse rerankers can be applied to perform event extraction with minimum domain specific training.  They use an off-the-shelf dependency parser, MSTParser \cite{McDonald2005,McDonald2006}, but extend it with event-specific features. Their approach requires conversion to and from dependency trees, at the beginning and and at the end. The features in the MSTParser are quite local (i.e., able to examine a portion of each event at a time); the decoding necessary can be performed globally, allowing the dependency parser some trade-offs. 
Event parsing is performed using three modules: 1) anchor detection to identify and label event anchors, 2) event parsing to form candidate event structures by linking entries and event anchors, and 3) event reranking to select the best candidate event structure. First, they parse the sentences with a reranking parser \cite{Charniak2005} with the biomedical parsing model from  \cite{McClosky2010}, using the set of Stanford dependencies \cite{deMarneffe2008}. After the parsing, they perform anchor detection using a technique inspired by techniques for named entity recogntion to label each token with an event type or {\em none}, using a logistic regression classifier. The classifier uses features inspired by \cite{Bjorne2009}. They change a parameter to obtain high recall to overgenerate event anchors. Multiword event anchors are reduced to their syntactic head. The event anchors and the included entities become a ``reduced" sentence, input to the event parser. Thus, the event parser gets words that are believed to directly take part in the events. This stage uses the MSTParser  with additional event parsing features. The dependency trees are decoded and  converted back to event structures. Finally, for event reranking, the system gets $n$ best list of event structures from each decoder in the previous step of event parsing. The reranker uses global features of an event structure to restore and output the highest scoring structure. The reranking approach is based on parse reranking \cite{Ratnaparkhi1999}, but is based on features of event structures instead of syntactic constituency structure. They use the  {\em cvlm} estimator \cite{Charniak2005} when learning weights for the reranking model. Since the reranker can work with outputs of multiple decoders, they use it as an ensemble technique as in \cite{Johnson2010}. 

The FAUST system \cite{Riedel2011a} shows that using a straightforward model combination strategy with two competitive systems, the UMass system \cite{Riedel2011b} and the Stanford system \cite{McClosky2011} just described,  can produce a new system with substantially high accuracy. The new system uses the framework of {\em stacking} \cite[Chapter 17]{Alpaydin2010}. 
The new system does it by including the predictions of the Stanford system into the UMass system, simply as a feature. 
 Using this simple model of  stacking, the FAUST system was able to obtain first place in three tasks out of four where it participated.

\begin{figure}
	\centering
	\includegraphics[scale=0.6]{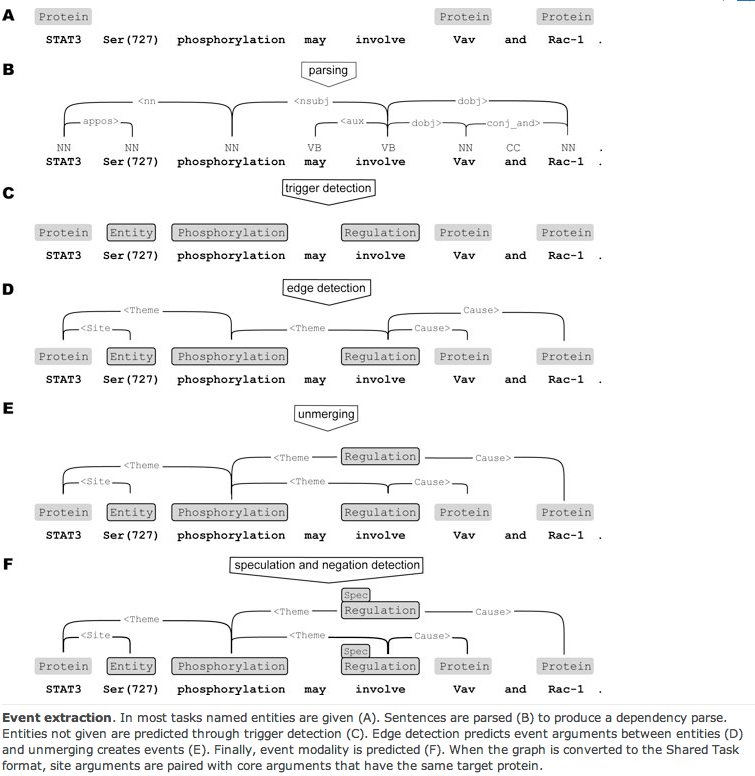}
	\caption{Turku BioNLP Pipeline}
	\label{fig:Turku2011System}
\end{figure}

The Turku Event Extraction System \cite{Bjorne2011,Bjorne2012} can be easily adapted to different event schemes, following the theme of event generalization in BioNLP 2011. The system took part in eight tasks in BioNLP 2011 and demonstrated the best performance in four of them. 
The Turku system divides event extraction into three main steps: i) Perform named entity recognition in the sentence, ii) Predict argument relations between entities, and iii) Finally, separate entity/argument sets into individual events. The Turku system uses a graph notation with trigger and protein/gene entities as nodes and relations (e.g., theme) as edges. In particular, an event in the graph representation is a trigger node along with its outgoing edges. The steps are shown in Figure \ref{fig:Turku2011System}. 
The Turku system uses Support Vector Machines \cite{Vapnik1995,Tsochantaridis2006} at various stages to perform each of the sub-tasks.   
 To use an SVM classifier, one needs to convert text into features understood by the classifier. The Turku system performs a number of analyses  on the sentences, to obtain features, which are mostly binary. The features are categorized into token features (e.g., Porter-stem \cite{Porter1980}, Penn Treebank part-of-speech tags \cite{Marcus1993}, character bi- and tri-grams, presence of punctuation on numeric characters), sentence features (e.g., the number of named entities in the sentence), dependency chains (up to a depth of three, to define the context of the words), dependency with $n$-grams (joining a token with two flanking dependencies as well as each dependency with two flanking tokens), trigger features (e.g.,  the trigger word a gene or a protein) and external features (e.g., Wordnet hypernyms, the presence of a word in a list of key terms). 
Applicable combinations of these features are then used by the three steps in event detection: trigger detection, edge detection and unmerging. Trigger words are detected by classifying each token as negative or as one of the positive trigger classes using SVMs. Sometimes several triggers overlap, in which case a merged class (e.g. {\em phosphorylation--regulation}) is used. After trigger prediction, triggers of merged classes are split into their component classes. Edge detection is used to predict event arguments or triggerless events and relations, all of which are defined as edges in the graph representation. The edge detector defines one example per direction for each pair of entities in the sentence, and uses the SVM classifier to classify the examples as negatives or as belonging to one of the positive classes.  When edges are predicted between these nodes, the result is a merged graph where overlapping events are merged into a single node and its set of outgoing edges. To produce the final events, these merged nodes need to be Òpulled apartÓ into valid trigger and argument combinations. Unmerging is also performed using the SVM classifier. Speculation and negation are detected independently, with binary classification of trigger nodes using SVMs. The features used are mostly the same as for trigger detection, with the addition of a list of speculation-related words.

\section{Extracting Events from Socially Generated Documents}
With the explosive expansion of the Internet during the past twenty years, the volume of socially generated text has skyrocketed. Socially generated text includes blogs and microblogs.  
For example, Twitter\footnote{\url{http://www.twitter.com}}, started in 2006, has become a social phenomenon. It allows individuals with accounts to post short messages that are up to 140 characters long.  Currently, more than 340 million tweets are sent out every day\footnote{\url{http://blog.twitter.com/2012/03/twitter-turns-six.htm}}. While a majority of posts are  conversational or not particularly meaningful, about 3.6\% of the posts concern topics of mainstream news\footnote{\url{http://www.pearanalytics.com/blog/tag/twitter/}}. Twitter has been credited with providing the most current news about many important events before traditional media, such as the attacks in Mumbai in November 2008. Twitter also played a prominent role in the unfolding of the troubles in Iran in  2009 subsequent to a disputed election, and the so-called Twitter Revolutions\footnote{\url{http://en.wikipedia.org/wiki/Twitter_Revolution}} in Tunisia and Egypt in 2010-11. 

Most early work on event extraction of information from documents found on the Internet has focussed on news articles \cite{Chambers2011,Doddington2004,Gabrilovich2004}. However, as noted earlier, social networking sites such as Twitter and Facebook have become important complimentary sources of such information. 
Individual tweets, like SMS messages, are usually short and self-contained and therefore are not composed of complex discourse structures as is the case with texts containing narratives. 
However, extracting structured representation of events from short or informal texts is also challenging because most tweets are about mundane things, without any news value and of interest only to the immediate social network. Individual tweets are also very terse, without much context or content. In addition, since Twitter users can talk about any topic, it is not clear a priori what event types may be appropriate for extraction.

\begin{figure}
	\centering
	\includegraphics[scale=0.6]{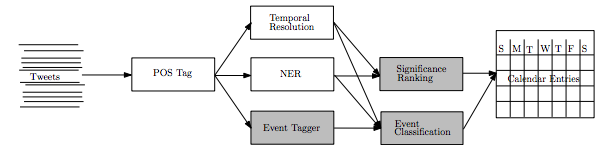}
	\caption{TwiCal Architecture}
	\label{fig:TwiCalArchitecture}
\end{figure}

The architecture of the system called TwiCal for event extraction \cite{Ritter2012} from Twitter messages is given in Figure \ref{fig:TwiCalArchitecture}. 
Given a stream of raw tweets, TwiCal extract events with associated named entities and times of occurrence. First the tweets are POS tagged using a tagger \cite{Ritter2012}, especially trained with Twitter data. Then named entities are recognized \cite{Ritter2011} using a recognizer trained with Twitter data as well.  After this, phrases that mention events (or,   event triggers or event phrases or just events)  are extracted using supervised learning.   \cite{Ritter2012} annotated 1,000 tweets with event phrases, following guidelines for annotation of EVENT tags in Timebank \cite{Pustejovsky2003}. 
The system recognizes event triggers   as a sequence labeling task
using Conditional Random Fields \cite{Lafferty2001}. It uses a contextual dictionary, orthographic features, features based on the Twitter-tuned POS tagger, and dictionaries of event terms gathered from WordNet \cite{Sauri2005}.  Once a large number of events have been extracted by this CRF learner, TwiCal  categorizes these events into types using an unsupervised approach based on latent variable models, inspired by work on modeling selectional preferences \cite{Ritter2010,Seaghdha2010,Kozareva2010,Roberts2011} and unsupervised information extraction \cite{Bejan2009,Chambers2011,Yao2011}. This automatic discovery of event types is similar to topic modeling, where one automatically identifies the extant topics in a corpus of text documents. 
The automatically discovered types (topics) are quickly inspected by a human effort to filter out incoherent ones, and the rest are annotated with informative labels. Examples of event types discovered along with top  event phrases and top  entities are given in Table \ref{table:TwiCalEventTypes}.   The resulting set of types are applied to categorize millions of extracted events  without the use of any manually annotated examples. 
 For inference, the system uses collapsed Gibbs sampling \cite{Griffiths2004} and prediction is performed using a streaming approach to inference \cite{Yao2009}.  To resolve temporal expressions, TwiCal uses TempEx \cite{Mani2000}, which takes as input a reference date, some text and POS tags, and marks temporal expressions with unambiguous calendar references.
Finally, the system measures the strength of association between each named entity and date based on the number of tweets they co-occur in, in order to determine if the event is significant. 
Examples of events extracted by TwiCal are given in Table \ref{table:TwiCalExamples}. Each event is a 4-tuple including a named entity, event phrase, calendar date and event type. 

\begin{table}
\begin{center}
\begin{tabular}{| l | p{1.95in} |  p{1.95in} | } \hline 
Label & Top  Event phrases & Top Entities \\ \hline \hline 
Sports & tailgate, scrimmage, tailgating, homecoming & espn, ncaa, tigers, eagles\\ \hline
Concert & concert, presale, performs, tickets & taylor swift, toronto, britney spears, rihanna \\ \hline
Perform & matinee, musical, priscilla, wicked & shrek, les mis, lee evans, broadway \\ \hline
TV & new season, season finale, finished season, episodes & jersey shore, true blood, glee, dvr, hbo\\ \hline
\end{tabular}
\end{center}
\caption{Some of Examples of event types extracted by TwiCal}
\label{table:TwiCalEventTypes}
\end{table}

\begin{table}
\begin{center}
\begin{tabular}{| l | l | l | l |} \hline 
Entity & Event phrase & Date & Type \\ \hline \hline 
Steve Jobs & died & 10/6/11 & Death\\ \hline
iPhone & announcement & 10/4/11 & ProductLaunch \\ \hline
GOP & debate & 9/7/11 & PoliticalEvent \\ \hline
Amanda Knox & verdict & 10/3/11 & Trial\\ \hline
\end{tabular}
\end{center}
\caption{Examples of events extracted by TwiCal}
\label{table:TwiCalExamples}
\end{table}

The TwiCal system describe above used topic modeling using latent variables as one of the several computational components; it is used to capture events  captured using supervised learning into types or topics. 
\cite{Weng2011} point out some drawbacks of using such an approach.  The main problem is that frequently the result generated by Latent Dirichlet Analysis (LDA) is difficult to interpret because it simply gives a list of words associate with the topic. For example, when \cite{Weng2011} attempt to find the four most important
 topics using LDA based on a Twitter collection emanating from Singapore on June 16, 2010, they find the topics listed in Table \ref{table:SingaporeLDATopics}. 
Therefore,  Weng et al. present another approach to detect events from a corpus of Twitter messages. Their focus is on detection and therefore, not on extraction of components that describe an event. Event detection is based on the assumption that when an event is taking place, some related words show an increase in usage. In this scheme, an event is represented by a number of keywords showing a burst in appearance count \cite{Yang1998,Kleinberg2003}. 
Although it is clear that tweets report events, but such reports are usually overwhelmed by high flood of meaningless ``babbles". In addition, the algorithms for event detection must be scalable to handle the torrent of Twitter posts. The EDCoW (Event Detection with Clustering of Wavelet-based Signals) system builds signals for individual words by applying wavelet analysis \cite{} on frequency-based raw signals of words occurring in the Twitter posts. These signals capture only the bursts in the words' appearance. The signals are computed efficiently by wavelet analysis \cite{Kaiser2011,Daubechies1992}.  Wavelets are quickly vanishing oscillating functions and unlike sine and cosine functions used in Discrete Fourier Transformation (DFT) \cite{}which are localized in frequency but extend infinitely in time, wavelets are localized both in time and frequency. Therefore, wavelet transformation is able to provide precise measurements about when and to what extent bursts take place in a signal. \cite{Weng2011} claim that this makes it a better choice for event detection when building signals for individual words. Wavelet transformation converts signals from time domain to time-scale domain where scale can be considered the inverse of frequency.  Such signals also take less space for storage. Thus, the first thing EDCoW does is convert frequencies over time to wavelets, using a sliding window interval.  It removes trivial words by examining signal auto-correlations. The remaining words are then clustered to form events with a modularity-based graph partitioning technique, which uses a scalable eigenvalue algorithm. 
It detects events by grouping sets of words with similar patterns of burst. To cluster, similarities between words need to be computed. It does so by using cross correlation, which is a common measure of similarity between two signals \cite{Orfanidis1985}. Cross correlation is a pairwise operation. Cross correlation values among a number of signals can be represented in terms of a correlation matrix $\mathcal M$, which happens to be a symmetric sparse matrix of adjacent similarities. With this graph setup, event detection can be formulated as a graph partitioning problem, i.e., to cut the graph into subgraphs. Each subgraph corresponds to an event, which contains a set of words with high cross correlation, and also that the cross correlation between words in different subgraphs are low. The quality of such partitioning is measures using a metric called modularity \cite{Newman2004,Newman2006}. The modularity of a graph is defined as the sum of weights of all the edges that fall within subgraphs (after partitioning) subtracted by the expected edge weight sum if the edges were placed at random. The main computation task in this component is finding the largest eigenvalue and corresponding eigenvector, of the sparse symmetric modularity matrix. This is solved using power iteration, which is able to scale up with the increase in the number of words in the tweets \cite{Ipsen2006}. 
EDCoW requires each individual event to contain at least two words. 
To differentiate big events from trivial ones, EDCoW quantifies the events' significance, which depends on two factors, the number of words and cross-correlation among the words related to the event. To make EDCoW work with TwiCal to see if it improves performance, the topic detection module will have to be replaced. EDCoW associates fewer words to topics because it filters words away before associating with a topic. Table \ref{table:EDCoWEvents} gives a few event words obtained by EDCoW and the corresponding event description. Please note that the event description was created by the authors and not the system.

\begin{table}
\begin{center}
\begin{tabular}{| l | l |} \hline 
Topic ID & Top Words  \\ \hline \hline 
13 & flood, orchard, rain, spain, road, weather, singapor, love, cold\\ \hline
48 & time, don, feel, sleep, love, tomorrow, happy, home, hate \\ \hline
11 & time, love, don, feel, wait, watch, singapor, hope, life \\ \hline
8 & watch, world, cup, match, time, love, don, south, goal\\ \hline
\end{tabular}
\end{center}
\caption{Examples of Topics Detected by LDA from Singapore based tweets on June 16, 2010}
\label{table:SingaporeLDATopics}
\end{table}

\begin{table}
\begin{center}
\begin{tabular}{| l | l |} \hline 
Event Words & Event Description \\ \hline \hline 
democrat, naoto & Ruling Democratic Party of Japan elected Naoto Kan as chief \\ \hline
ss501, juju & Korean popular bands Super Junior's and SS501's performance on mubank \\ \hline
\#kor, greece, \#gre & A match between South Korea and Greece in World Cup 2010l\\ \hline
\end{tabular}
\end{center}
\caption{Examples of Events Detected by EDCoW in June 2010}
\label{table:EDCoWEvents}
\end{table}

\subsection{Summarization}
\cite{Filatova2004} use event-based features to represent sentences and shows that their approach improves the quality of the final summaries compared to a baseline bag-of-words approach. 

\subsection{Question Answering}
Event recognition is a core task in question-answering since the majority of web questions have been found to be relate to events and situations in the world \cite{Sauri2005}. For example, to answer the question {\em How many people were killed in Baghdad in March?}, or {\em Who was the Prime MInister of India in when China and India fought  their only war?}, the question-answering system may have to identify  events across a bunch of documents before creating an answer.

\section{Future Directions of Research}

It also seems like when doctors take notes on a patient's history or medical record, the information is not written in order of events or in temporal order all the time. It will be good to take notes from here and there and put them in an event ordered fashion or temporally ordered manner. Extracting an event based structure of the medical record would help understand  the medical history better. 

Most systems process sentences in isolation, like most event extraction systems at the current time. Therefore, events crossing sentence boundaries cannot be detected.

\bibliographystyle{acmtrans}
\bibliography{events}

\begin{received}
...
\end{received}
\end{document}